\documentclass{article}
\usepackage{ijcai22}

\usepackage{times}
\usepackage{soul}
\usepackage{helvet}  
\usepackage{courier}  
\usepackage[hyphens]{url}  
\usepackage[hidelinks]{hyperref}
\usepackage[utf8]{inputenc}
\usepackage[small]{caption}
\usepackage{graphicx}
\usepackage{amsmath}
\usepackage{amsthm}
\usepackage{booktabs}
\usepackage{caption} 
\usepackage{algorithm}
\usepackage{algorithmic}

\usepackage[normalem]{ulem}
\usepackage{amssymb}
\usepackage{mathtools}
\usepackage{comment}
\usepackage{ulem}

\usepackage{subcaption}
\usepackage[para,online,flushleft]{threeparttable}
\usepackage{multirow}
\usepackage{dcolumn}
\usepackage{booktabs}
\usepackage{makecell}

\usepackage{comment}
\usepackage{enumitem}
\setlist[itemize]{noitemsep, topsep=0pt, label=$\blacktriangleright$, leftmargin=*}
\newtheorem{theorem}{Theorem}
\usepackage{color}
\usepackage{xspace}
\usepackage[textsize=tiny,color=gray!40]{todonotes}

\newtheorem{myprop}{\bf{Proposition}}

\newtheorem{myremark}{\bf{Remark}}

\newtheorem{statement}{Statement}

\newcommand{\argmin}{\operatornamewithlimits{arg\,min}}

\DeclarePairedDelimiterX{\inp}[2]{\langle}{\rangle}{#1, #2}
\newcommand{\bx}{\mathbf{x}}

\newcommand{\bh}{\mathbf{h}}
\newcommand{\bu}{\mathbf{u}}
\newcommand{\bg}{\mathbf{g}}
\newcommand{\din}{\mathcal D^{\texttt{tr}}}
\newcommand{\dout}{\mathcal D^{\texttt{val}}}
\newcommand{\btheta}{\boldsymbol{\theta}}
\newcommand{\bphi}{\boldsymbol{\phi}}

\newcommand{\grad}{\nabla}

\newcommand{\task}{\mathcal{T}}
\newcommand{\rnn}{\text{{\tt RNN}}\xspace}

\newcommand{\E}{\mathbb{E}}

\newcommand{\R}{\mathbb{R}}

\newcommand{\pphi}[1]{\frac{\partial #1}{\partial \bphi}}

\DeclareMathOperator*{\minimize}{\text{minimize}}

\DeclareMathOperator*{\stt}{\text{subject to}}
\DeclareMathAlphabet\mathbfcal{OMS}{cmsy}{b}{n}
\newcommand{\Def}[0]{\mathrel{\mathop:}=}
\newcommand{\highlightmath}[2]{\colorbox{#1!50}{$\displaystyle#2$}}

\renewcommand{\paragraph}[1]{\noindent{\bf #1}}
\usepackage{diagbox} 
\usepackage{multirow}
\newcommand{\high}[1]{\textcolor{black}{\textbf{#1}}}
\usepackage{wrapfig}
\usepackage[rightcaption]{sidecap}
\usepackage{adjustbox}

\newcommand{\mycomment}[1]{}

\newcommand{\SL}[1]{\textcolor{blue}{SL: #1}}

\newcommand{\lft}{LFT\xspace}

\title{Learning to Generate Image Source-Agnostic Universal Adversarial Perturbations}

\author{
Pu Zhao$^1$
\and
Parikshit Ram$^2$\and
Songtao Lu$^2$\and
Yuguang Yao$^3$\and\\
Djallel Bouneffouf$^2$\and
Xue Lin$^1$\And
Sijia Liu$^{2,3}$
\affiliations
$^1$Northeastern University, \quad
$^2$IBM Research, \quad 
$^3$ Michigan State University
\emails
zhao.pu@northeastern.edu, \{parikshit.ram, songtao\}@ibm.com, yaoyugua@msu.edu, djallel.bouneffouf@ibm.com, xue.lin@northeastern.edu, liusiji5@msu.edu
}

\begin{document}

\maketitle

\begin{abstract}
% Adversarial perturbations are critical for training robust deep learning models. 
{Adversarial perturbations are critical for certifying the robustness of deep learning models.}
%Often, adversarial perturbations for a given victim model are generated separately for each image. An alternate approach is the generation of
%Besides instance-wise adversarial perturbations, 
A ``universal adversarial perturbation'' (UAP)  can simultaneously attack multiple images, and thus offers a more unified threat model, 
% without calling for 
{obviating}
an image-wise attack  algorithm. However, the existing UAP generator is underdeveloped when images are drawn from different image sources (e.g., with different image resolutions). Towards an \textit{authentic universality across image sources},  we %formulate  
 take a novel
 view of 
UAP generation   as a customized 
 instance of
% ``meta-learning'',
{``few-shot learning''}, which  
 leverages bilevel optimization and learning-to-optimize (L2O) techniques for UAP generation with improved attack success rate (ASR). 
% To be specific, w
We begin by considering the popular model agnostic meta-learning (MAML) framework to meta-learn a UAP generator.
%over just few-shot image classification tasks. 
However, we see that the MAML framework does not directly offer the  universal attack across image sources,  requiring us to integrate it with another meta-learning framework of L2O. The resulting scheme for meta-learning a UAP generator (i) has better performance (50\% higher ASR) than baselines such as Projected Gradient Descent, % that does not leverage meta-learning, 
(ii) has better performance (37\% faster) than the vanilla L2O and MAML frameworks (when applicable), 
and (iii) is able to simultaneously handle UAP generation for different victim models and  image data sources.
\end{abstract}

\section{Introduction} \label{sec:intro}
Adversarial perturbations are imperceptible changes to input examples (such as images) aimed at manipulating the predictions of a ``victim model''~\cite{madry2017towards,carlini2017towards}. These are essential for evaluating the worst-case robustness of deep learning (DL) models, which is critical when deploying such models to real world scenarios. The usual focus in adversarial attack generation is on the perturbation of an individual data sample for a single victim model~\cite{madry2017towards}. A more powerful threat model is that of {\em universal adversarial perturbation} (UAP)~\cite{moosavi2017universal} to simultaneously perturb multiple examples. This is often accomplished with standard optimizations  such as Projected Gradient Descent (PGD)~\cite{madry2017towards}, optimizing for a {\em single} perturbation that simultaneously attacks a set of provided examples.

\paragraph{Motivation.}
Although  UAP has been widely studied in the literature 
%various attack generation methods are proposed to construct UAPs 
\cite{li2020regional,khrulkov2018art,liu2019universal}, three 
%\todo{Is there a 3rd one we need to add?} 
fundamental \textbf{challenges} (\hyperlink{c1}{\texttt{C1}}-\hyperlink{c3}{\texttt{C3}}) remain. 

(\hypertarget{c1}{\texttt{C1}}) \textit{Lack of generalization ability to unseen images}:  
%Given that the UAP is generated using a set of examples, it is reasonable to expect that same perturbation to be able to attack other previously unseen examples from the same distribution on the same victim model to some degree~\cite{moosavi2017universal}. However, existing UAP generation schemes 
UAP can attack other previously unseen examples from the same distribution on the same victim model to some degree~\cite{moosavi2017universal}. However, existing UAP methods are usually unable to  attack unseen images with high success rate 
% \textcolor{blue}{when trained with limited seen examples}
when {generated} with limited seen examples.
% (
% %see  Figure~\ref{fig:pgd-uap} in 
% \S \ref{sec:uap-fs}).
%\todo{Skip ref to \S 3}.
%\todo{We need to compare ASR of PGD on train vs. unseen test}

%\textcolor{blue}{
(\hypertarget{c2}{\texttt{C2}}) \textit{Less effective on new images}:  
For a new set of examples, the UAP usually needs to be regenerated by rerunning  the generation algorithm, which is less effective. It is desirable to develop more effective UAP optimization algorithms   on new data examples within one- or few-step updates. % } 

(\hypertarget{c3}{\texttt{C3}}) \textit{Non-applicability to diverse image sources}.
Beyond existing work,
% \SL{[1) did not follow the following sentence. We want to say disadvantage of PGD-UAP or advantage? I thought it should be disadvantage. 2) I  make a new paragraph to introduce new questions.]}
%However,
a more advantageous and authentic UAP generation scheme should be
%\todo{I was trying to say that PGD-UAP does not perform well on unseen images but can be applied to any dataset/victim model since it is a generic optimizer. So we want to keep that capability in our solution, motivating Q(ii)}
%However, one significant advantage of such PGD-based UAP generation schemes is that they are 
broadly applicable to any image source and corresponding victim model -- the same attack generator %(in this case, an optimizer) 
can simultaneously generate perturbations for images   from different data sources (with different  resolutions).

% (\texttt{C3}) \textbf{High computation complexity of UAP generation at test time}. 
% Compared to the generation of adversarial attack with respect to (w.r.t.) a single image, the existing UAP generator becomes more computationally-intensive when optimizing over  multiple images. Thus, how to scale up UAP generation at test time is another challenge.

\paragraph{Research idea and rationale.}
The    challenges \hyperlink{c1}{\texttt{C1}}-\hyperlink{c3}{\texttt{C3}}  bring us to the central question  we aim    to answer in this paper:

%\textcolor{blue}{
{\em (Q) ``Can we develop a powerful  threat model in the form of a UAP generator that  can improve the attack performance on unseen examples {with a small set of seen examples} (\hyperlink{c1}{\texttt{C1}}), and  generate UAPs more effectively within a few steps (\hyperlink{c2}{\texttt{C2}}) for images from different sources (\hyperlink{c3}{\texttt{C3}})?'' 
}
%}
%and  simultaneously generate UAP attacks for images   from different sources (\hyperlink{c2}{\texttt{C2}})?''}
%\todo{Can we also say ``different victim models''?}
%\todo{\textcolor{red}{I think we can. There are some UAP works  investigating the transferibility of UAPs across different model architectures. }  
%}

To address (\hyperlink{c1}{\texttt{C1}}) and (\hyperlink{c2}{\texttt{C2}}), 
%To answer the first question \textit{(i)}, 
we formulate  %a novel interpretation of 
the UAP generation process as an instance of {\em few-shot learning}, and explore the use of \textit{model-agnostic meta-learning} (MAML) 
%standard 
%meta-learning~\citep{vanschoren2018meta} 
techniques \cite{finn2017model}  
to warm-start the learning when only few examples are available. We show that
MAML 
%equips the UAP generator with a generalized universality as it 
enables us to  learn a good meta-model of UAP
%in the context of learning with neural networks. It focuses on gradient-based learning and meta-learns 
%an initialization of a neural network 
%(for supervised and reinforcement learning)
with an \textit{explicit goal of fast adaptation} -- the ability to quickly learn  
% transferable
{generalizable}
UAP 
with just a few examples. 
We highlight that this is  different from existing computationally-intensive UAP generators, requiring to perturb a large volume of images for improved 
attack generalizability.

To address (\hyperlink{c3}{\texttt{C3}}),  namely, accomplish a source-agnostic UAP model, we present an extension of MAML to `\textit{incongruous} tasks' -- tasks drawn from diverse image sources -- by leveraging
a different meta-learning framework, {\em learning-to-optimize} (L2O)~\cite{li2017learning,andrychowicz2016learning}.
The conventional MAML only applies   to `\textit{congruous} tasks' where  the meta-learning and task-specific learning (fine-tuning) occur on the same set of learnt parameters. %(parameters that are optimized over during the learning process).
{In UAP generation, the task-specific parameters are the image perturbations, whose size depend on the image sizes.}
%By contrast, t
To remain agnostic to the image source and size, it is not possible to share the learnt parameters
% (the UAP in this case) 
between different image sources. Thus, MAML is not directly applicable. 
%That is, the tasks in our  setup are {\em incongruous}, not allowing us to directly apply MAML to our problem. 
% To tackle this, we employ L2O to learn UAP  over incongruous tasks. 
To tackle this, we employ L2O to {meta-learn the UAP generator} over incongruous tasks. 
Our rationale is that L2O provides us with a \textit{learnt optimizer},
%e.g., in terms of a trained  model rather than hand-crafted optimizers such as  SGD. 
% These learnt optimizers 
which use gradients or zeroth-order gradient estimates~\cite{liu2020primer},
and
can operate on objectives with \textit{different set of optimizee variables} {(task-specific parameters)}, %.
%This allows meta-learning across incongruous tasks.
allowing meta-learning across incongruous tasks.

We highlight that MAML and L2O make complementary contributions to  UAP: 
% MAML focuses on meta-learning ``where to start learning''; 
MAML focuses on meta-learning {``for better generalization''};
L2O meta-learns ``how to learn''. 

\paragraph{Contributions.} We outline our contributions as follows:  ({\S} refers to section number):

\begin{itemize}
\item (\S \ref{sec:uap-fs}) 
We propose a novel interpretation of UAP threat model as a few-shot learning problem.
%meta-learning problem.
%through the lens of bilevel optimization. %allowing us to leverage MAML and L2O.  allowing us to leverage the rich literature of meta-learning.
\item (\S\ref{sec:lft})
Algorithm-wise,
we show how meta-learning across incongruous tasks can be developed using  an extended bilevel optimization by integrating L2O with MAML and applied to  UAP generation. Theory-wise, we quantify how our  meta-learned fine-tuner (\lft) differs from L2O.
\item (\S\ref{sec:emp-eval}) 
We demonstrate the improved attack performance of our  meta-learning based UAP generator against standard optimization based generators and other meta-learning based UAP generators (when applicable in limited scope).
\end{itemize}

\paragraph{Notation.}
In the few-shot learning setup, we denote the $i^{\text{\tt th}}$ individual task as $\task_i \sim P(\task)$ sampled from a task distribution $P(\task)$ with the task specific 
(i)~learning parameters (or {\em optimizee variables}) $\btheta_i \in \Theta_i$,
(ii)~data domain $D_i$, 
% (iii)~support (training or fine-tuning) set $\din_i \in D_i$,
(iii)~support (training/fine-tuning/seen) set $\din_i \in D_i$,
% (iv)~query (validation or test) set $\dout_i \in D_i$,
(iv)~query (validation/test/unseen) set $\dout_i \in D_i$,
(v)~objective function $f_i: \Theta_i \times D_i \to \R$. 
In the MAML framework, the tasks are congruous, and share the optimizee variables, hence the optimizee domains are the same, that is, $\Theta_i = \Theta_j = \Theta \ \ \forall i,j$, and a single $\btheta \in \Theta$ is meta-learned and fine-tuned to $\btheta_i \in \Theta$ for any task $\task_i$. In the L2O framework, the tasks can be incongruous, and only share the optimizer parameters $\bphi$ while maintaining their own separate {\em optimizee variables} $\btheta_i$. %We will elaborate on this further in \S \ref{sec:lft}.

\section{Related work} \label{sec:related}
%\vspace*{-0.00in}
%\SL{[add important references to Introduction, and reduce the length.]}
%

\paragraph{Adversarial attacks and UAP.}
There   exist an extensive amount of work on the design of adversarial attacks, ranging from white-box attacks to black-box attacks \cite{carlini2017towards,ruan2020learning,Goodfellow2015explaining,papernot2016cleverhans,zhao2019design,chen2017zoo,xu2018structured}. 
In the context of UAP, various attack generation methods were proposed \cite{li2020regional,khrulkov2018art,liu2019universal,hashemi2020transferable,matachana2020robustness}. For example, \citeauthor{li2020regional} proposed regionally homogeneous perturbations.
\citeauthor{khrulkov2018art} leveraged the  singular vectors of the Jacobian matrices of deep features to construct UAP. \citeauthor{liu2019universal} designed a robust UAP generation by fully exploiting the model uncertainty. 
% \textcolor{blue}{
% \citeauthor{yuan2022meta}  design a better initialization for  causing  misclassifications  after being updated through a one-step gradient ascent update.\todo{Not UAP}}
To boost the attack generalizability, most UAP  methods require to perturb a large volume  of images simultaneously %, making UAP generation  computationally-intensive at test time. 
However, this makes UAP generation computationally-intensive at test time.
Moreover, existing UAP generation is restricted to a single image source, leaving image source-agnostic UAP an open question. %In this work, we address the above  problems (namely, \hyperlink{c1}{\texttt{C1}}-\hyperlink{c3}{\texttt{C3}} in  \S \ref{sec:intro}) by leveraging  MAML for few-shot learning and L2O.  

\paragraph{MAML.}
 MAML~\cite{finn2017model} has been extremely useful in supervised and reinforcement learning (RL), and widely extended and studied both theoretically~\cite{liu2019taming,balcan2019provable,khodak2019adaptive} and empirically~\cite{nichol2018first-order}. It has been extended to model uncertainty~\cite{finn2018probabilistic} and handle the online setting~\cite{finn2019online}. 
 %\todo{here ...}
 The second order derivatives in MAML have been handled in multiple ways~\cite{rajeswaran2019meta,fallah2019convergence}. Specific to RL, various enhancements obviate the second order derivatives of the RL reward function, such as variance reduced policy gradients~\cite{liu2019taming} and Monte Carlo zeroth-order Evolution Strategies gradients~\cite{song2020esmaml}.
 %\todo{..to here can remove}
 %\cite{mutlikai20} present a Hessian-free MAML with multiple fine-tuning steps. \cite{yao2019hierarchically} propose a hierarchically structured meta-learning to explicitly tailor transferable knowledge to different task clusters. 
 %The core idea is to perform cluster-specific meta-learning, leading to tighter generalization bounds. 
% \SL{MAML has also been applied to adversarial attack generation [xxxx].}
 However, all extensions and applications of MAML focus on congruous tasks where different few-shot tasks share the same parameters and optimizee domain.

\paragraph{L2O.}
Learnt optimizers have long been considered in   training neural networks~\cite{bengio1990learning,thrun2012learning}.
More recent work has posed optimization {\em with gradients} as a RL  problem~\cite{li2017learning} or as learning a recurrent neural network (\rnn)~\cite{andrychowicz2016learning} instead of leveraging the usual hand-crafted optimizers (such as SGD or  Adam~\cite{KingmaB2015adam}). The \rnn based optimizers have been improved
~\cite{wichrowska2017learned,chen2020training} by -- (i)~using hierarchical {\rnn}s to better capture parameter structure of DL models, (ii)~using hand-crafted-optimizer-inspired inputs to \rnn (such as momentum), (iii)~using a diverse set of optimization objectives (with different hardness levels) to meta-learn the \rnn, and (iv)~leveraging different training techniques such as curriculum learning and imitation learning. The learnt optimizers have also been successful with particle swarm optimization~\cite{cao2019learning} and zeroth-order gradient estimates~\cite{ruan2020learning}. Learnt optimizers have been used for meta-learning with congruous few-shot tasks~\cite{ravi2016optimization}, but MAML has been shown to outperform it.
In the context of adversarial robustness, 
L2O has also been leveraged  to design instance-wise attack generation \cite{ruan2020learning,jiang2018learning} and  adversarial defense (such as adversarial training \cite{xiong2020improved}).

\section{UAP Design as a Few-Shot Problem} \label{sec:uap-fs}

%crafted to manipulate the DNN prediction.
%Design of adversarial attacks has attracted a significant amount of research interests as the crafted adversarial examples provide a natural way to evaluate the worst-case robustness of a learned DNN \cite{Goodfellow2015explaining}.

%Though adversarial attacks has attracted many research interests, 
Many existing attack generators execute in an instance-wise manner, namely, 
requesting repeated invocations of an iterative optimizer to acquire adversarial perturbations with respect to an \textit{individual} input example \cite{chen2017zoo,croce2020reliable}.
To circumvent the limitation of instance-wise attack generator, 
the problem of  UAP arises, which seeks a \textit{single} perturbation pattern
to  manipulate the DNN outputs   over \textit{multiple examples simultaneously} 
\cite{moosavi2017universal,matachana2020robustness}.

\paragraph{Problem setup.}
Let $\task_i$ denote an attack generation task, which 
constitutes a support set $\din_i$ used to generate the UAP $\btheta_i$ and a query set $\dout_i$ (from the same classes as in $\din_i$) on which we evaluate the ASR (attack success rate) of the generated UAP $\btheta_i$.
Our \textit{goal} is to learn a UAP with a few examples in $\din_i$ that can successfully attack \textit{unseen} examples $\dout_i$  with the same victim model.
%This positioning as few-shot learning allows us to tap into the rich literature of meta-learning. 
This motivates us to view UAP generation as a {\bf few-shot learning} problem. % \citep{finn2017model}.
Mathematically,
given a few-shot UAP generation task $\task_i$ with training set $\din_i$, the task-specific UAP $\btheta_i$ is obtained by solving: % the following:

\vspace*{-3mm}
{\small
\begin{align}\label{eq:cw_loss}
%  \begin{array}{ll}
\hspace*{-2mm} \displaystyle \minimize_{\btheta_i} \ 
  f_{i}(\btheta_i, \din_i)       \Def
  \displaystyle \sum_{(\bx, y) \in \din_i}
  \ell_{\mathrm{atk}}(\btheta_i, \mathbf x, y)
  % \max \{    p(  \bx + \btheta )_{y} - \max_{c \neq y}  \{  p(\bx + \btheta )_c\}, 0 \}
 + \lambda \| \btheta_i \|_1,
%    \end{array}
\hspace*{-3mm}
\end{align}}%
%\todo{Can we say that \eqref{eq:cw_loss} is just one example of attack loss but that we can work with any attack loss function? }%
%\todo{\textcolor{red}{ yes, we can use other loss. But CW loss is the best loss function compared with a number of other loss functions. That is why it is widely used. } }%
where $y$ is the true label of $\bx$, $\lambda > 0$ is a regularization parameter, and $\ell_{\mathrm{atk}}$ is the C\&W attack loss~\cite{carlini2017towards}, which is $0$ (indicating a successful attack) with an incorrect predicted class.
%when the incorrect class is predicted as the top-1 class. 
The second term % of \eqref{eq:cw_loss} 
is an $\ell_1$ norm regularizer,  penalizing the perturbation strength.  %of  $\btheta$. %, measured by its $\ell_1$ norm.  
%Note that another type of attack loss different from \eqref{eq:cw_loss} can also be used in our proposed UAP generator. For simplicity, we focus on the C\&W attack loss. 

In our few-shot notation, the attack loss \eqref{eq:cw_loss} is considered as the task-specific loss $f_i$ with $\btheta_i$ as the task-specific parameters. With multiple few-shots tasks $\task_i$ and corresponding $(\din_i, \dout_i)$, we wish to meta-learn a UAP generator that can solve new few-shot UAP generation tasks.
%Note that each few-shot task can be drawn from different data sources.

To achieve this, we leverage 
MAML to meta-learn an initialization of {\em optimizee} variables $\btheta$ (i.e., UAP variables)
%for few-shot learning 
that enables fast adaptation to new tasks when fine-tuning the optimizee from this learned initialization with   a few new examples.
%~\citep{finn2017model,finn2019online}.
%\todo{see what we can shrink given existing notation discussion. SL: I think this is good.}
Formally, with $N$ few-shot learning tasks $\{ \task_i \}_{i=1}^N$, when meta-learning with  $\task_i$, {\it (i)}~the support set $\din_i$ is used for the task-specific {\em inner level} in MAML to fine-tune the initial optimizee $\btheta$, and {\it (ii)}~the query set $\dout_i$ is used in the {\em outer level} for  evaluating the fine-tuned optimizee $\btheta_i^*(\btheta)$ to meta-update % the initialization 
$\btheta$.
Thus, \textbf{MAML-oriented UAP generation} solves the following {\bf bilevel optimization problem:}
%\todo{Should we highlight that this is novel application of MAML for UAP gen?}
{\small
\vspace{-0.02mm}
\begin{align}
\label{eq:maml-orig}
\begin{array}{l}
\displaystyle \minimize_{\btheta}      \displaystyle \frac{1}{N}\sum_{i = 1}^N \E_{(\din_i, \dout_i) \sim \task_i} \left [ f_{i} ( \btheta_i^*(\btheta); \dout_i)
\right ], \\ 
\stt \ \ %\mathbf z^* (\btheta)
\btheta_i^*(\btheta)
= \displaystyle \argmin_{\btheta_i} f_i(\btheta_i(\btheta); \din_i) %, \btheta),
\end{array}
\vspace{-0.02mm}
\end{align}}%
where
$\btheta_i(\btheta)$ is the task-specific optimizee and
%$f_i(\btheta_i; \mathcal D, \btheta)$
$f_i(\btheta_i (\btheta); \mathcal{D})$
is the task-specific loss evaluated on data $\mathcal D$ using variable $\btheta_i(\btheta)$ obtained from fine-tuning the meta-learned initialization $\btheta$.
%Problem \eqref{eq:maml-orig} provides a generalized formulation of MAML.
% Note that problem \eqref{eq:maml-orig} is a bilevel optimization problem since  $\btheta_i^*(\btheta)$  is actually a function of $\btheta$. 
We will use $\btheta_i:= \btheta_i(\btheta)$ from hereon.

% \begin{figure}[htb]
% \vspace{-1mm}
% \centering{
% \hspace*{-2mm}
% \includegraphics[width=0.25\textwidth]{figs/mcasr_pgd.pdf}
% }
% \vspace*{-3mm}
% \caption{ \footnotesize{UAP attack success rate vs. \# PGD steps  on $\din_i$ (seen images) and $\dout_i$ (unseen images) for CIFAR-10.
% }}
% \vspace*{-3mm}
% \label{fig:pgd-uap}
% \end{figure}

\paragraph{Why   \eqref{eq:maml-orig}? Formulation \eqref{eq:cw_loss}   is not able to generate transferable UAP.}
%In what follows, 
We provide a warm-up example showing why we choose to rely on the bilevel few-shot   formulation \eqref{eq:maml-orig}, rather than \eqref{eq:cw_loss}. 
We thus examine if solving the task-specific   problem \eqref{eq:cw_loss} is sufficient to generate UAP with 
``attack generalizability'' when applied to unseen test example.
%Figure~\ref{fig:pgd-uap} demonstrates a numerical example where 
We consider a UAP generation  on CIFAR-10 and use  PGD to generate the task-specific UAP $\btheta_i$ over 1000 tasks {each with 4 images}.
%\todo{Is it possible to have a similar plot/line for S-UAP in Figure~\ref{fig:pgd-uap}?}
We compare the average ASR of $\btheta_i$ on $\din_i$ (with which the UAP was generated) and $\dout_i$ (unseen images from same classes). %, aggregated over all the tasks. 
Since the UAP $\btheta_i$ is generated with $\din_i$, the ASR achieves 100\%  on $\din_i$. However, the same $\btheta_i$ when applied to $\dout_i$ achieves {\em less than 30\% ASR} -- this highlights how the PGD-generated UAP that only solves problem \eqref{eq:cw_loss}   has low ASR on unseen images from the same distribution on the same victim model. Formulation~\eqref{eq:maml-orig} explicitly minimizes the loss on unseen examples in the outer-loop, thereby explicitly promoting attack generalizability, allowing us to address \hyperlink{c1}{\texttt{C1}} in  \S \ref{sec:intro}.
%transferable to unseen test examples. 

 %\begin{wrapfigure}[23]{l}[0pt]{0.42\textwidth}
 \begin{figure}  %[htb]
 %\vspace{-5mm}
    \centering
    \includegraphics[width=0.43\textwidth]{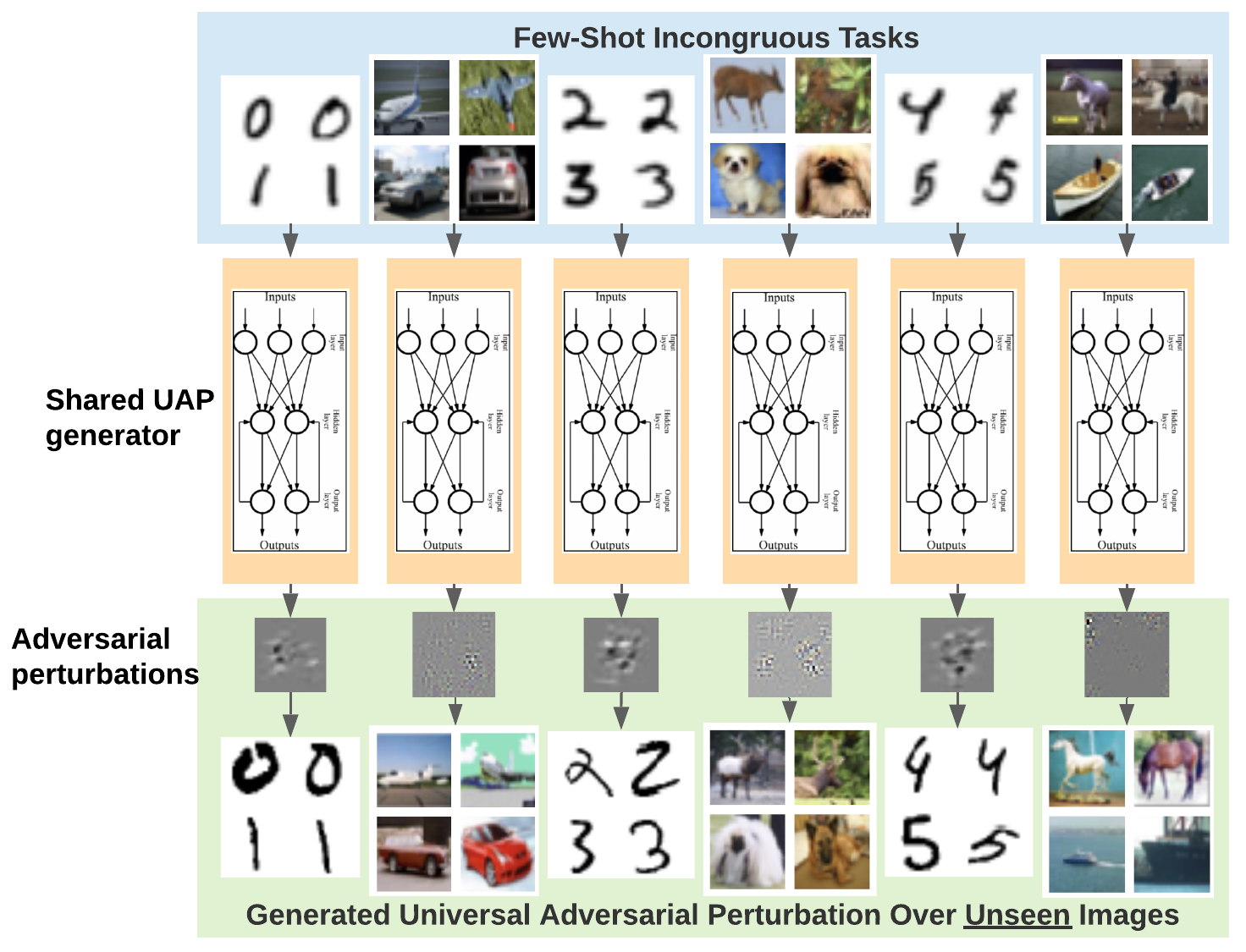}
%     \vspace*{-5mm}
    \caption{
    {Desired UAP generation setup} with {\em incongruous few-shot}  tasks across $28\times28$ MNIST and $3 \times 32 \times 32$ CIFAR-10. The meta-learned   optimizer parameters  are shared by all incongruous  tasks to find task-specific UAP patterns even with various image sizes. % images have different sizes across tasks. 
    % For few-shot tasks based on MNIST and CIFAR-10, we use the LeNet-5~\citep{lecun2015lenet} architecture based models. The meta-learning of the shared \rnn optimizer $\bphi$ uses the task-specific meta-gradients. Note that MAML cannot be applied in this incongruous setting to meta-learn a UAP parameter initialization for both tasks in MNIST and CIFAR-10.
    % 
    }
    \vspace{-10pt}
    \label{fig:UAPgen-viz}
\end{figure}

%\end{wrapfigure}

% Second, as mentioned earlier, we want the UAP generator to possess a {\bf generalized universality} across hybrid image sources. The UAP generator should be able to (i) solve the few-shot task of designing a UAP with a few ``training'' images that can successfully attack (generalize attacks to) unseen ``test'' images, and (ii)  solve few-shot tasks (that is, design attacks) with images for different data sources -- {\em incongruous few-shot tasks}. We visualize this in Figure~\ref{fig:UAPgen-viz}
% %\todo{maybe highlight what is UAP generator, what are the perturbations. maybe change ``meta-learn shared ... optimizer'' to ``shared UAP generator''} 
% where the {\em same UAP generator}
% (which will show to be a {\rnn}-based learnt optimizer) can solve few-shot tasks from MNIST \cite{lecun1998mnist} % (with CC BY-SA 3.0 license)
% as well as CIFAR-10 \cite{Krizhevsky2009learning}. % (with MIT license).

\section{Learning Optimizers for Fast Adaptation} \label{sec:lft}
% \section{Problem Formulation} \label{sec:prob-form}
%
While \eqref{eq:maml-orig}  improves attack generalizability and addresses challenges \hyperlink{c1}{\texttt{C1}} and \hyperlink{c2}{\texttt{C2}} stated in {Sec.}\,\ref{sec:intro}, it is only applicable to `congruous tasks' where all UAP generation tasks are from the same image source.  %for the images from the same source.
In this section, 
we generalize \eqref{eq:maml-orig} to `\textbf{incongruous tasks}' with learned fine-tuners ({\lft}s) by integrating with  L2O, thereby addressing \hyperlink{c3}{\texttt{C3}} to  yield image source-agnostic UAP generators.

\paragraph{MAML and beyond.}
To solve  problem \eqref{eq:maml-orig},  the conventional approach~\cite{finn2017model,Yin2020Meta-Learning} relies on the approximation of 
the inner problem   with a $K$-step gradient descent (GD) with the initial iterate 
$\btheta_i^{(0)} \gets \btheta$, the final iterate
$\btheta_i^* \gets \btheta_i^{(K)}$ and

\vspace*{-3.8mm}
{\small
%\vspace*{-0.2in}
\begin{align}\label{eq: inner_GD}
%    \btheta_i^* &=  \btheta_i^{(K)}, \\
    \btheta_i^{(k)} &= \btheta_i^{(k-1)}- \alpha \nabla_{\btheta_i}
    f_i(\btheta_i^{(k-1)}; \din_i) %, \btheta)
    \text{ for }
    k  \in [K],%  \\  \btheta_i^{(0)} &= \btheta,
\end{align}
}%
where $\btheta_i^{(k)}$ is the $k^{\text{th}}$-step optimizee fine-tuned with $\din_i$ from  $ \btheta_i^{(0)} = \btheta$, $\alpha > 0$ is a learning rate,
and $[K] = \{ 1,2,\ldots, K\}$. 
% Although GD \eqref{eq: inner_GD} and variants solve the inner minimization in \eqref{eq:maml-orig} efficiently, the outer loop requires
% the second-order derivative with respect to (w.r.t.) $\btheta$.
% With large $K$, MAML faces the issue of vanishing gradients.

It is clear from \eqref{eq: inner_GD} that both levels of the optimization (with respect to $\btheta$ and $\btheta_i$) in \eqref{eq:maml-orig} must
 operate on the {same-type} optimizee variables, %(i.e., perturbation variables of the same dimension),
and accordingly, few-shot tasks $\{ \task_i\}_{i=1}^N$ are restricted to problems
%domains with the \textit{same} dimension, namely,
%$\mathrm{dim}(\btheta_i^*) = \mathrm{dim}(\btheta_j^*) $ for   $\forall i,j \in [N]$.
which {\em share the same optimizee domain} (i.e., \textit{congruous} tasks drawn from the same image source).
%For example, all learning tasks are required to share the same network model parameters.
However, in the general meta-learning setting, similar tasks could be from related yet \textit{incongruous} domains corresponding to \textit{different objectives} $\{ f_i \}$ with optimizee variables of \textit{different domains} (such as different image sizes and number of channels) that \underline{cannot be shared between tasks} as in our UAP generation tasks illustrated in Figure~\ref{fig:UAPgen-viz} -- 
UAP parameters cannot be shared between images from different data sources with different resolutions.
%Network parameters from different architectures cannot be (generally) shared even if they are solving related learning tasks.
%networks of different sizes being used to solve similar problems.
%
In such  cases, meta-learning the \textit{initial iterate is not possible}. Next, we will  leverage L2O to  meta-learn an optimizer -- the fine-tuner --
%rather than  initialization
for fast adaptation of the task-specific optimizee $\btheta_i$ in a few-shot setting even when meta-learning across \textit{incongruous} tasks.

%Spurred by the
\paragraph{L2O-enabled learned fine-tuner for MAML.}
L2O allows us to replace the  hand-designed GD \eqref{eq: inner_GD}  with
a learnable \rnn parameterized by $\bphi$.
%Specifically, the
For any task $\task_i$,
%the model 
$\rnn_{\bphi} (\cdot)$ mimics a hand-crafted gradient based optimizer
%the rule of  optimization, which
to output a descent direction $\Delta \btheta_i$
to update task-specific optimizee variable $\btheta_i$ given the function gradients as input. Thus, we replace \eqref{eq: inner_GD}  with
{\small
\vspace*{-0.04in}
\begin{align}\label{eq: RNN_update}
  \begin{array}{l}
       \Delta \btheta_i^{(k)}, \bh_i^{(k)} = \rnn_{\bphi} \left (
g_i(\btheta_i^{(k-1)}; \din_i), \bh_i^{(k-1)} \right),    \\  
         \btheta_i^{(k)} = \btheta_i^{(k-1)} - \Delta \btheta_i^{(k)}, \quad \forall k \in [K],
 \end{array}
\end{align}
}%
where $\bh_i^{(k)}$ denotes the state of $\rnn_{\bphi}$ at the $k^{\text{th}}$ \rnn unrolling step,
%$\btheta_{\task, 0}$ is a given initialization,
$g_i(\btheta_i^{(k)}; \din_i)$ is the gradient $\grad_{\btheta_i}
f_i (\btheta_i; \din_i)$ or gradient estimate ~\cite{liu2017zeroth}.
Each task-specific $\btheta_i$ is initialized with a random $\btheta_i^{(0)}$. 
Note that $\Delta \btheta_i^{(k)} := \Delta \btheta_i^{(k)}(\bphi)$ is a function  of $\bphi$ in \eqref{eq: RNN_update} and hence $\btheta_i^{(k)} := \btheta_i^{(k)}(\bphi)$  depends on  $\bphi$.

We term $\rnn_{\bphi}$ in \eqref{eq: RNN_update} {\em learned fine-tuner} (\lft)  for incongruous few-shot learning.
Combining  \eqref{eq: RNN_update} with \eqref{eq:maml-orig}, we can cast the meta-learning of a \lft as 
{\small
\vspace*{-0.06in}
\begin{align}
\hspace*{-3mm} \label{eq:LFT}
\begin{array}{l}
\displaystyle \minimize_{\bphi} ~  \displaystyle \underbrace{ \frac{1}{N}\sum_{i = 1}^N \E_{(\din_i, \dout_i) \sim \task_i} \sum_{k=1}^K   w_k f_i ( \btheta_i^{(k)}(\bphi); \dout_i) }_\text{$:= \widehat{F}(\bphi)$}\\
\stt ~  
\text{$\btheta_i^{(k)}(\bphi)$ is given by  \eqref{eq: RNN_update}},
\end{array}
\end{align} }%
where
$w_k$ is an importance weight for the $k^{\text{th}}$ unrolled step in \eqref{eq: RNN_update}.
We can set 
(i)~$w_k = 1$~\cite{andrychowicz2016learning}, 
(ii)~$w_k = k$~\cite{ruan2020learning}, or 
%(iii) the indicator function w.r.t. last unrolled step 
(iii)~$w_k = \mathbb{I}[k = K]$~\cite{lv2017learning}. Choice (iii) matches the  objective \eqref{eq:maml-orig}, focusing on the final fine-tuned solution. However, unlike MAML, problem \eqref{eq:LFT} meta-learns the {\em fine-tuner} $\bphi$ instead of an initialization $\btheta$. 
We elide the $\bphi$ argument in the sequel for brevity.

The fine-tuner $\bphi$ acquired  from   \eqref{eq:LFT} yields
%Second, as mentioned earlier, we want 
a  \textit{UAP generator} with {generalized universality} across hybrid image sources. First, it can generate  a UAP $\btheta_i^{(k)}(\bphi)$ with a few ``seen'' images in $\din_i$  that can successfully attack ``unseen'' images in $\dout_i$ (\hyperlink{c1}{\texttt{C1}}) within a few updating steps (\hyperlink{c2}{\texttt{C2}}). Second, as will be evident later, its form given by a {\rnn}-based  optimizer enables us to handle incongruous few-shot tasks (that is, design attacks) with images for different data sources (\hyperlink{c3}{\texttt{C3}}).
\begin{algorithm}[tb]
%\begin{algorithm}[H]
\caption{Meta-learning \lft %$\rnn_{\bphi}$ (the UAP generator)
with problem \eqref{eq:LFT}}
\label{alg:LFT}
%{\small
\begin{footnotesize}
\begin{algorithmic}[1]
\STATE {\bf Input:} UAP generation tasks $\{ \task_i \}_{i \in [N]}$,
%learning rate $\beta$,
%number of  iterations $T$, number of fine-tuning steps $K$,
\# meta-learning steps $T$, \# fine-tuning steps per task $K$,
initial $\bphi$, $\{ \btheta_i^{(0)}, \bh_i^{(0)}  \}_{i \in [N]}$,
meta-learning rate $\beta > 0$
\FOR{$t \gets 1,2,\ldots,T$}
  %\STATE sample batch of tasks $\mathcal B_t \subseteq [N]$
  \FOR{tasks $i \in $ sampled task batch $\mathcal{B}_t \subseteq N$}
    \STATE sample data $\din_i \sim \task_i$ for fine-tuning
    \STATE \textit{generate task-specific optimizee (the UAP) with $\din_i$ via \eqref{eq: RNN_update}} for $K$ steps to obtain $\{ \btheta_i^{(k)}, \bh_i^{(k)} \}_{k=1}^K$
    \STATE sample data $\dout_i \sim \task_i$ for the meta-update
    \STATE obtain task-specific {\em fast-adaptation gradient} w.r.t. \lft parameters $\bphi$ with $\dout_i$
      {\small
      \vspace{-6pt}
      \begin{align}\label{eq: grad_outer}
      \bg^{(i)} \gets \grad_{\bphi}
 \textstyle     \sum_{k=1}^K  \left [ w_k f_i ( \btheta_i^{(k)}; \dout_i)  \right ]
      \end{align}}%
    %\STATE
    %save task-specific state
    %  $\btheta_i^{(0)}, \bh_i^{(0)} \gets \btheta_i^{(K)}, \bh_i^{(K)}$
    %\todo[inline, author=PR]{@Sijia, @Pu is this line 8 necessary? Technically we always start with random initial $\btheta_i^{(0)}$ for any task during meta-testing.}
      %\Comment{For EI, save all $f_i$ evaluations}
  \ENDFOR
  \STATE \textit{update \lft parameters $\bphi$:}
  $
    \bphi \gets \bphi - \beta \sum_{i \in \mathcal B_t} \bg^{(i)}
   $
   \COMMENT{// Adam can also be used}
\ENDFOR
\STATE {\bf Output:} $\rnn_{\bphi}$
\end{algorithmic}
\end{footnotesize}
\end{algorithm}
%\end{wrapfigure}
%
\subsection{Methodologies} \label{sec:lft:alg-frwrk}
The meta-learning problem~\eqref{eq:LFT} is still a bilevel optimization, similar to MAML~\eqref{eq:maml-orig}. However, both inner and outer levels are distinct from MAML:
In the inner level, we update a task-specific optimizee $\btheta_i$ by unrolling $\rnn_{\bphi}$ for $K$ steps from a \textit{random} initial state $\btheta_i^{(0)}$; by contrast, MAML uses GD to update $\btheta_i$ from the meta-learned initialization. % (that is, $\btheta_i^{(0)}\gets \btheta$).
In the outer level, we minimize the objective \eqref{eq:LFT} w.r.t. the optimizer $\bphi$ instead of the optimizee initialization $\btheta$. We present our proposed scheme in Algorithm\,\ref{alg:LFT}. 
In each outer iteration $t \in [T]$ of the meta-learning, we sample a batch $B_t$ of few-shot UAP generation tasks (line 3), and for each sampled task $\task_i, i \in B_t$, we obtain a {set of seen images} $\din_i$ (line 4) and use it with the \lft to generate a (sequence of) UAP $\btheta_i^{(k)}$ using the update rule in \eqref{eq: RNN_update}. Then we sample a set of unseen images $\dout_i$ (line 6) and generate the {\em fast adaptation gradient} w.r.t. $\bphi$ (line 7) that explicitly takes into account the utility of the generated UAP $\btheta_i^{(k)}$ on unseen images. These gradients w.r.t. $\bphi$ are aggregated across all tasks in the batch and used to update the \lft parameters $\bphi$ (line 9).
%
% \SL{It is worth mentioning that although the proposed meta-learning (Alg.~\ref{alg:LFT}) enjoys a similar
% meta-learning framework as  \citep{ravi2016optimization} which  also employed L2O to conduct task-specific fine-tuning, the motivation is significantly different: We aim to tackle incongruous few-shot learning tasks rather than congruous tasks considered in \citep{ravi2016optimization}. Moreover, we provide  in-depth analysis of  Alg.~\ref{alg:LFT},
% }
In what follows, we discuss our proposed meta-learning (Alg.~\ref{alg:LFT}),
showcasing its
%(i) generalizability to meta-learning across incongruous tasks,
(i) ability to meta-learn across incongruous tasks,
(ii) applicability to zeroth-order (ZO) optimization,
(iii) theoretical advantage from the fast adaptation gradient~\eqref{eq: grad_outer}.

\paragraph{Incongruous meta-learning.}
When fine-tuning the task-specific optimizee $\btheta_i$ by $\rnn_{\bphi}$ (Algorithm\,\ref{alg:LFT}, Step~5), we use an \textit{invariant} \rnn architecture to tolerate the task-specific variations in the domains $\{\Theta_i\}_{i=1}^N$ (e.g., dimensions) of optimizee variables $\{ \btheta_i \}_{i=1}^N$. Recall from \eqref{eq: RNN_update}  that $\rnn_{\bphi}$ uses the gradient or gradient estimate $g_i(\btheta_i^{(k)}; \din_i)$ as an input, which has the same dimension as $\btheta_i$. At first glance, a single $\rnn_{\bphi}$ seems incapable of handling {incongruous} $\{ \task_i \}_{i=1}^N$ defined over optimizee variables of different dimensionalities. However, a $\rnn_{\bphi}$ configured as a \textit{coordinate-wise} Long Short Term Memory
(LSTM) network \cite{andrychowicz2016learning}), is \textit{invariant} to the dimensionality of optimizee variables $\{ \btheta_i \}_{i=1}^N$ 
% That is because it independently operates on each coordinate of  $\btheta_i $ regardless of its dimensionality.
{by using a separate LSTM to independently operate on each coordinate of $\btheta_i$ for any task $i \in [N]$, but requiring all LSTMs to {\em share their weights}. This weight-sharing allows the \rnn to operate on tasks with optimizee variables of different dimensionalities.}
In contrast to MAML, the invariant $\rnn_{\bphi}$ expands meta-learning for fast adaptation beyond congruous tasks to incongruous ones such as designing UAPs across incongruous attack tasks. 

\paragraph{Derivative-free meta-learning.}
L2O in \eqref{eq: RNN_update} allows us to update the task-specific optimizee variable $\btheta_i$ using not only first-order (FO) information (gradients) but also zeroth-order (ZO) information (function values) if the loss function $f_i$ is a black-box objective function.
%In this case, one 
We can estimate the gradient $\grad f_i (\btheta_i; \mathcal{D}_i)$ with finite-differences of function values $g_i(\btheta_i; \mathcal{D}_i)$~\cite{liu2020primer,ruan2020learning}:
{\small
%\vspace*{-0.05in}
\begin{align} \label{eq:egrad}
g_i(\btheta_i; \mathcal{D}_i)  %\nonumber \\
=  %\frac{1}{\mu n}
   \frac{
   \sum_{j = 1}^n \left [ \bu_j \left(
      f_i(\btheta_i + \mu \bu_j; \mathcal{D}_i)
      - f_i(\btheta_i ;\mathcal{D}_i)
   \right) \right ]
   }{\mu n},
\end{align}
}%
where $\mu > 0$ is a small smoothing parameter, $\bu_j, j \in [n]$ are $n$ random directions with entries from $\mathcal{N}(0,1)$.
%the normal distribution.
%, and $n$ is the number of random directions.
% This gives us ZO-\lft alongside our original FO-LFT.
The function $g_i$
%$g_i(\cdot; \cdot)$ 
can also be sophisticated quantities derived from gradients or gradient estimates~\cite{wichrowska2017learned,lv2017learning,cao2019learning}.
%This gives us ZO-\lft along with the original FO-LFT.
%
The support for ZO optimization is crucial when   explicit gradient  are computationally difficult or infeasible in the black-box attack setting.
%, as is the case in UAP generation tasks for black-box adversarial attacks.
%of the gradients are difficult {to compute} or infeasible to obtain.
%In the application of universal adversarial perturbation generation,
%Our experiments show that ZO-\lft can be as effective as state-of-the-art FO-\lft, despite leveraging only the inputs and outputs of the  DL model (see application of designing black-box UAP in \S\ref{sec: app_atk}).
%
%\todo[author=PR,inline]{discussion on ZO-\lft and FO-\lft is too general; I don't think we have any evidence of this. \SL{ I think we compared ZO-\lft with FO-\lft in the last example; Fig. 3.}}

% \paragraph{Free of high-order derivative computation.}
%\paragraph{Recursive computation formulas.}
%\paragraph{Meta-learning Gradient.}
%
%
%\todo{here ...}
Since Alg.~\ref{alg:LFT} meta-learns the optimizer variable $\bphi$ rather than the  initialization of optimizee variable $\btheta$, it requires a different meta-learning gradient $\grad_{\bphi} \sum_k w_k f_i(\btheta_i^{(k)}, \dout_i)$.
Focusing only on $\nabla_{\bphi} f(\btheta_i^{(k)}; \dout_i)$ in \eqref{eq: grad_outer}:
{\small
\vspace*{-0.2mm}
\begin{equation}\label{eq: grad_meta_update}
\nabla_{\bphi} f_i(\btheta_i^{(k)}; \dout_i) = 
\underbrace{
  \frac{\partial  f_i(\btheta_i^{(k)}; \dout_i)}{\partial \btheta_i}
}_{g_i(\btheta_i^{(k)}; \dout_i)} \bullet %\cdot
\underbrace{
  \frac{\partial \btheta_i^{(k)}}{\partial \bphi}%  \nonumber \\
}_{\mathbf{G}_i^{(k)}}
\end{equation}
}%
where
%we drop the task specific index for ease of presentation,
% $\cdot$
$\bullet$ denotes a matrix product that the chain rule  obeys \cite{petersen2008matrix}. The computation of $\mathbf{G}_i^{(k)}$ calls for the first-order (or second-order) derivative of $f_i$ w.r.t. $\btheta_i$ if $g_i(\btheta_i^{(k)}; \din_i)$ denotes the gradient estimate %(or gradient respectively) 
of $f_i$ in \eqref{eq: RNN_update}. See Appendix\,\ref{asec:rnn-grad} for details\footnote{ A full version is available at https://arxiv.org/abs/2009.13714.}.

\paragraph{Theoretical advantage of fast adaptation gradient.}
%
%L2O is empirically very capable of meta-learning from optimization trajectories, with better convergence than hand-crafted optimizers~\citep{chen2020training}. Using notation from \S\ref{sec:prob-form}, the L2O objective can be written as:
We make use of the fast adaptation gradient \eqref{eq: RNN_update} explicitly evaluating the performance of the UAP (generated by fine tuning with $\din_i$) on unseen images in $\dout_i$. We want to highlight the advantage of this explicit choice -- we could have instead used the gradient of the fine-tuning objective $f_i(\btheta_i^{(k)}; \din_i)$ w.r.t. $\bphi$ by solving the following problem:

{\vspace*{-0.2in}
\small
\begin{align}\label{eq:L2O}
\begin{array}{l}
\displaystyle \minimize_{\bphi}    \displaystyle {F}(\bphi)=\frac{1}{N}\sum_{i = 1}^N \E_{\din_i \sim \task_i} \sum_{k=1}^K  w_k f_i ( \btheta_i^{(k)}; \din_i),   \\  
\mathrm{s.t.}  \  \    \text{$\btheta_i^{(k)}$ defined as \eqref{eq: RNN_update}}.
\end{array}
\end{align}}%
%\todo{make equation single line?}
%
Different from \eqref{eq:LFT}, this is   the standard L2O   in the context of few-shot tasks. Under mild assumptions, we show the following result, quantifying the difference between L2O and our proposed meta-learning in terms of the meta-learning gradient w.r.t. $\bphi$, highlighting the explicit effect of the fast adaptation gradient % in Algorithm~\ref{alg:LFT}
(see Appendix~\ref{asec:grad-diff-theory} for details):
%
%
%\vspace{-0.2cm}
\begin{myprop}\label{th.main}
Consider the meta-learning objectives defined in \eqref{eq:LFT} and \eqref{eq:L2O}. Suppose that the gradient size is bounded by $G$ and gradient estimate per sample has uniformly bounded variance $\sigma^2$. Then, for any $\bphi$, we have
%\vspace{-0.2cm}

{\small
\vspace*{-0.14in}
\begin{equation}
    \left\|\nabla_{\bphi} F(\bphi)-\nabla_{\bphi} \widehat{F}(\bphi)\right\|\leq \sqrt{2}G\sigma\sqrt{\frac{1}{D_{\texttt{tr}}}+\frac{1}{D_{\texttt{val}}}}
\end{equation}}%
where $D_{\texttt{tr}}:=\min_{i \in [N]} | \din_i | $ and $D_{\texttt{val}}:=\min_{i \in [N]} | \dout_i | $ denote the minimum batch size of per-task datasets $\din_i,\dout_i \sim \task_i, i \in [N]$.
%\SL{Here batch size means dataset size or small-batch size? the former?}
\end{myprop}
%\vspace{-0.2cm}
%\emph{Remark 1.}
%\begin{myremark}
%When  $D_{\texttt{tr}}$ and  $D_{\texttt{val}}$ are both large enough the difference between L2O and our proposed meta-learning is small. In this case, we reduce to L2O. From previous works, we know that L2O can meta-learn optimizers with good convergence properties, implying that our {\lft}s would also converge to similar results by leveraging the \rnn structure.
%\end{myremark}
%\emph{Remark 2.}
\begin{comment}
\begin{myremark}
When the data size is small -- the few-shot regime -- there could be a gap between the two frameworks, resulting in a significant difference in the solutions generated by L2O and \lft, %our scheme, 
especially for the case where $\sigma$ or $G$ is large.
%may result that the solution obtained by L2ML is sufficiently different from the one achieved by L2O.
%That would be reason why L2ML can outperform  L2O when the data samples are not enough, implying the superiority of the generalization power of L2ML compared with L2O.
This explains the significant difference between  empirical performance of L2O and our {\lft}s in  evaluation over few-shot learning problems in \S\ref{sec: app_atk}. %The effect of the batch sizes $\din_i, \dout_i$ on the difference in performance is presented in Appendix~\ref{appendix: classification}, validating the theory. 
%\SL{add this figure back.}
\end{myremark}
\end{comment}
\begin{myremark}
When the data size is small -- the few-shot regime -- there could be a significant difference in the gradients w.r.t $\bphi$, resulting in a significantly different solutions, especially for the case where $\sigma$ or $G$ is large.
This explains the significant difference between  empirical performance of L2O and our {\lft} in  evaluation over few-shot learning problems. 
%The effect of the batch sizes $\din_i, \dout_i$ on the difference in performance is presented in Appendix~\ref{appendix: classification}\todo{do we still keep these classification results?}, validating the theory. 
\end{myremark}

\vspace*{-2mm}
\section{Experiments} \label{sec:emp-eval}

We  demonstrate the effectiveness of \lft through extensive experiments on the UAP design  in the black-box  setup \cite{chen2017zoo}, where the internal  configurations and parameters of the DNN are not revealed to the attacker. Thus, the only interaction of the adversary with the system is via submission of inputs and receiving the corresponding predicted outputs. \lft is implemented using ZO gradient estimates.

\paragraph{Experimental setup.}
We wish to evaluate the ability of UAP generated with a small set of seen images to successfully attack unseen images from the same image source (for a fixed victim model). For this, we generate 100 UAP generation tasks $\task_i, i = 1, \ldots, 100$, each with a set $\din_i$ of seen images (to be used to generate the UAP) and a set $\dout_i$ of unseen images (on which the generated UAP is evaluated). In both $\din_i$ \& $\dout_i$, $2$ image classes with $2$ samples per class are randomly selected. As image sources, we utilize MNIST and CIFAR-10, and as victim architectures, we utilize LeNet~\cite{Lecun1998gradient} and VGG-11~\cite{simonyan2014very}. Note that the same architecture applied to different image sources will lead to different victim models.
For meta-learning, we separately generate an additional 1000 UAP generation tasks from each image source, ensuring that there is no overlap between images present in the evaluation (meta-test) and the meta-learning tasks.

\paragraph{Baselines.}
We consider two non-meta-learning based standard UAP generators -- PGD-based (termed PGD) and singular vector based (termed S-UAP)~\cite{khrulkov2018art}. In addition to our proposed meta-learning based LFT scheme, we also consider standard L2O (given by~\eqref{eq:L2O}) to ablate the effect of the fast adaptation gradient\footnote{Unseen images $\dout_i$ are not utilized in L2O~\eqref{eq:L2O}, so we add it to $\din_i$ to ensure that L2O gets the same images for meta-learning.}. We also consider standard MAML (given by~\eqref{eq:maml-orig}) to ablate the effect of using a meta-learnt optimizer (with random initial iterate) instead of a meta-learnt initialization. Note that MAML is only applicable if all the meta-learning and meta-testing UAP generation tasks come from a single image source. When the meta-learning tasks are generated from multiple image sources, we meta-learn a MAML per image source and term it {\em ensemble MAML}. Note that this requires us to solve the bilevel problem~\eqref{eq:maml-orig} separately for each image source.

\paragraph{Learnt optimizer details.}
We use a one-layer LSTM with 10 hidden units, and one additional linear layer to project the \rnn hidden state to the output. We use Adam with an initial learning rate of $0.001$ to meta-learn the \rnn with truncated backpropagation through time (BPTT) by unrolling the \rnn for 20 steps and running each optimization for 200 steps.

\paragraph{Evaluation metrics.}
We report (a)~averaged attack success rate  (ASR), (b)~$\ell_1$ perturbation strength, (c)~convergence in terms of optimization steps needed to reach 100\% ASR. We also provide visualization of generated UAPs in Appendix~\ref{appendix: l1norm}.

% RQ1: Does meta-learning lead to improved ASR on unseen images for few-shot UAP generation tasks?

\paragraph{Experimental questions and results.}
%We seek to answer the following questions with our experiments:

\begin{table}[t]
 \centering
  \caption{Attack success rate (ASR)  of UAPs generated by different schemes using 200 steps aggregated over 100 meta-test tasks from different image sources. We highlight the superior ASR of \lft.}
  \label{tab:rq1}
  \scalebox{0.74}[0.74]{
   \begin{threeparttable}
\vspace{-0.1in}
\begin{tabular}{c|c|c|c|c}
    \hline
\toprule[1pt]
\multicolumn{2}{c|}{\multirow{2}{*}{  \diagbox[width=10em,trim=l]{{\makecell{Meta-learning} }}{{\makecell{Meta-testing}}}  }} &  \multirow{2}{*}{ \makecell{MNIST \\ (LeNet-5)} }  &   \multirow{2}{*}{   \makecell{CIFAR-10 \\ (LeNet-5) }  }  & \multirow{2}{*}{\makecell{CIFAR-10 \\ (VGG-11) }}
\\
\multicolumn{2}{c|}{} &  &  \\
\midrule[1pt]
 \multirow {5}{*}{\makecell{MNIST \\ (LeNet-5)} } 
& \textcolor{black}{S-UAP} & 63 $\pm$ 5   &  42 $\pm$ 4 &  38 $\pm$ 4 \\ 
& PGD & 50$\pm$ 7 &  25 $\pm$ 0  &  25 $\pm$ 0  \\ 
& MAML & \high{100}$\pm$ 0 &  -  & - \\ 
& L2O &  \high{100}$\pm$ 0   &   40 $\pm$ 10  & 36$\pm$2   \\
%\hline
& \lft (Ours) & \high{100$\pm$} 0 &  \high{50 $\pm$} 0 & \high{48$\pm$} 3   \\
\bottomrule[1pt]
  \end{tabular}
\end{threeparttable}} \vspace{-10pt}
\end{table}

\begin{table*}[t]
 \centering
  \caption{Aggregated ASR and $\ell_1$-norm of UAPs  by various  schemes. We {highlight} the {best} performance at each (meta-learning and  -testing) scenario, measured by (i)~highest ASR within $100$ steps, (ii)~distortion at $100$ steps, and (iii)~steps required to first reach $100\%$ ASR. 
  }
  \label{tab:rq2}
\vspace{-0.1in}
  \scalebox{0.74}[0.74]{
   \begin{threeparttable}
\begin{tabular}{c|c|c|c|c|c|c|c|c|c|c}
    \hline
\toprule[1pt]
\multicolumn{2}{c|}{\multirow{2}{*}{  \diagbox[width=10em,trim=l]{{Meta-learning }}{{Meta-testing}}  }}
&  \multicolumn{3}{c|}{\texttt{T1}: MNIST}  &   \multicolumn{3}{c|}{\texttt{T2}: CIFAR-10} & \multicolumn{3}{c}{\texttt{T3}: MNIST + CIFAR-10}\\
\cline{3-11}
%\hline
%\midrule[1pt]
\multicolumn{2}{c|}{ } & 
\makecell{ASR$_{100}$(\%)}  & \makecell{$\ell_1$ norm}  & \makecell{step \#
%\tnote{1}
}
 & \makecell{ASR$_{100}$(\%)} &  \makecell{$\ell_1$ norm} & \makecell{step \#}   & \makecell{ASR$_{100}$(\%)} &  \makecell{$\ell_1$ norm} & \makecell{step \#}   \\
%\hline
 \midrule[1pt]
 \multirow {3}{*}{\makecell{\texttt{L2}:\\ MNIST + \\ CIFAR-10 }}
  & Ensemble MAML  &  75 $\pm$ 0 & 0.117  & 137  &  96 $\pm$ 4 & 0.071  &  88 &    95 $\pm$ 6 & 0.099  & 103 \\
 & L2O &  77 $\pm$ 8 & 0.112 & 125  &   95 $\pm$ 5 & 0.069  &  72 &  92 $\pm$ 9 & 0.096  & 93 \\
 &  \lft & \high{100 $\pm$} 0 & \high{0.101} & \high{92} &    \high{100 $\pm$} 0 & \high{0.063}  & \high{55}  & \high{100 $\pm$ }0 &  \high{0.089}  & \high{68}  \\
\bottomrule[1pt]
  \end{tabular}
%   \begin{tablenotes}
% \item[1] Step \# denotes the number of  steps when the average ASR first reaches 100\% within $200$ steps.
% \end{tablenotes}
\end{threeparttable}} %\vspace{-15pt}
\vspace{-0.1in}
\end{table*}

\begin{figure*}[tb]
 \centering
   \begin{adjustbox}{max width=1\textwidth }
\begin{tabular}{p{1.2in}p{1.4in}p{1.4in}p{1.4in}}
\toprule
\multicolumn{1}{c|}{
\diagbox[width=10em,trim=l]{{Meta-learning}}{{Meta-testing}}}
& \makecell{\centering {\footnotesize \texttt{T1}: MNIST}}  
& \makecell{\centering {\footnotesize \texttt{T2}: CIFAR-10}}  
& \makecell{\centering {\footnotesize \texttt{T3}: MNIST+CIFAR-10}}
\\
\midrule
\multicolumn{1}{c|}{\parbox[t][-1.2in][c]{0.4in}{ \centering \small{\texttt{L1}: MNIST} }}%{\footnotesize (Congruous tasks)} }
 &   %\hspace*{-0.1in}
\includegraphics[width=1.5in]{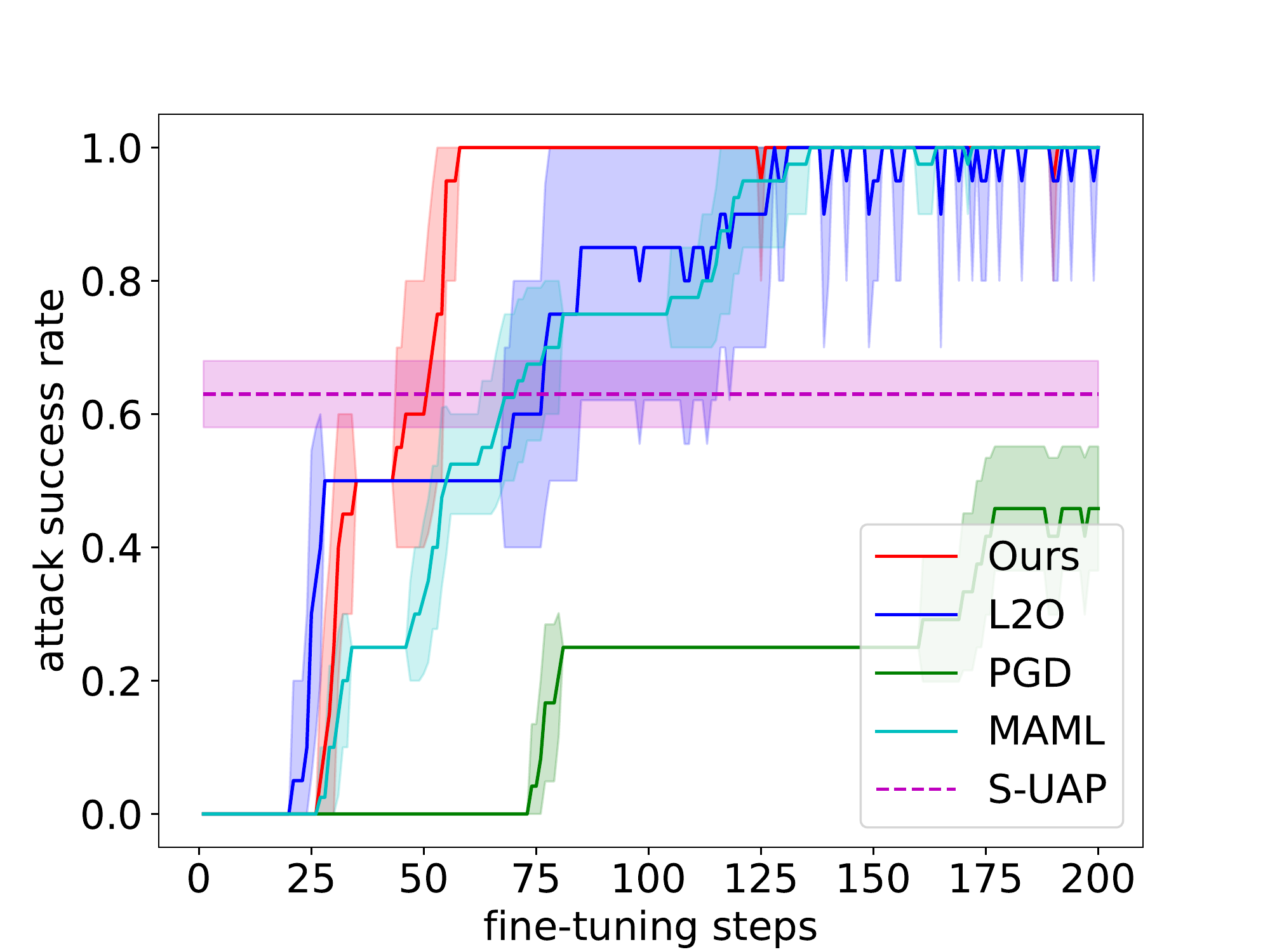}& %\hspace*{-0.1in}
\includegraphics[width=1.5in]{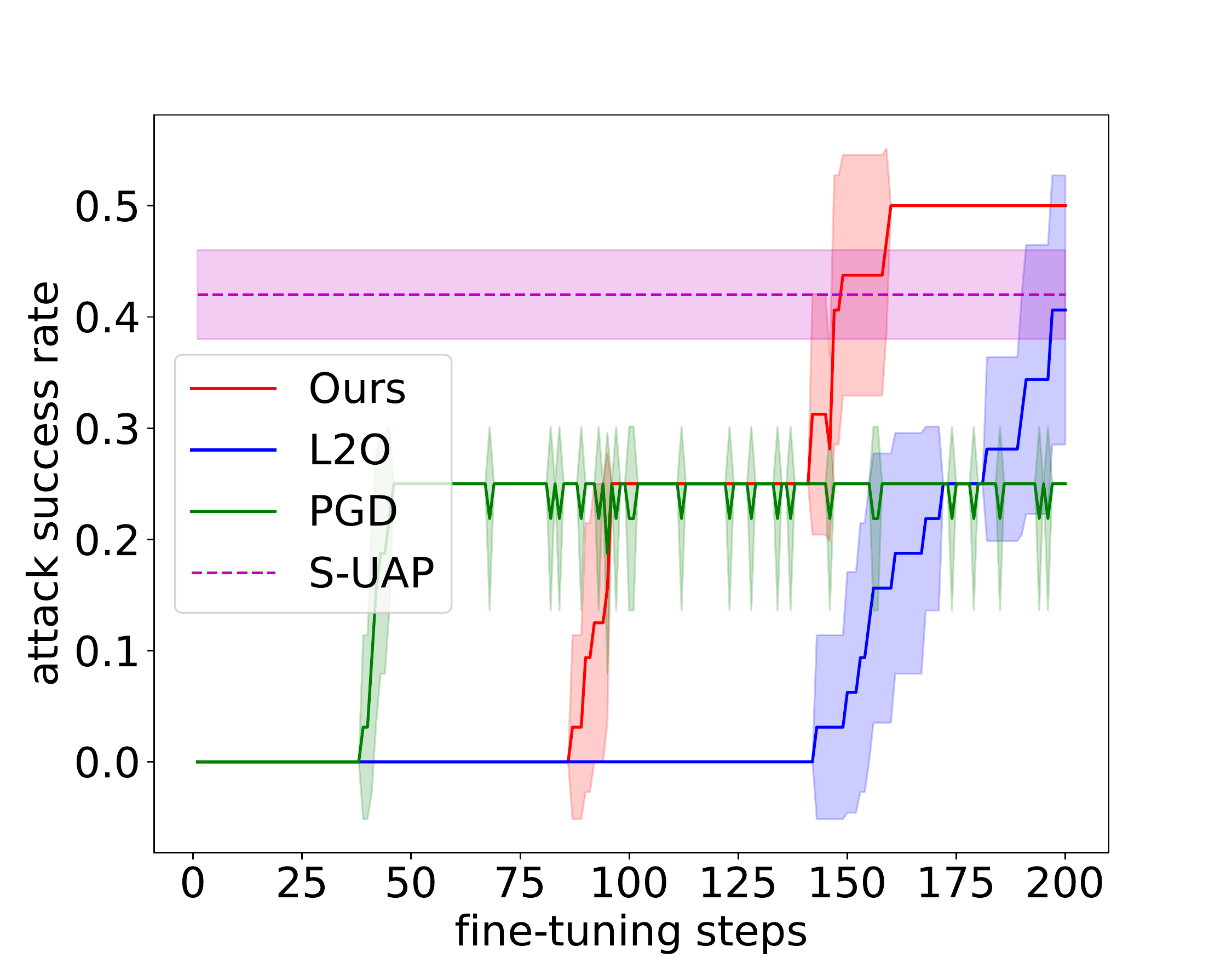}&
\includegraphics[width=1.5in]{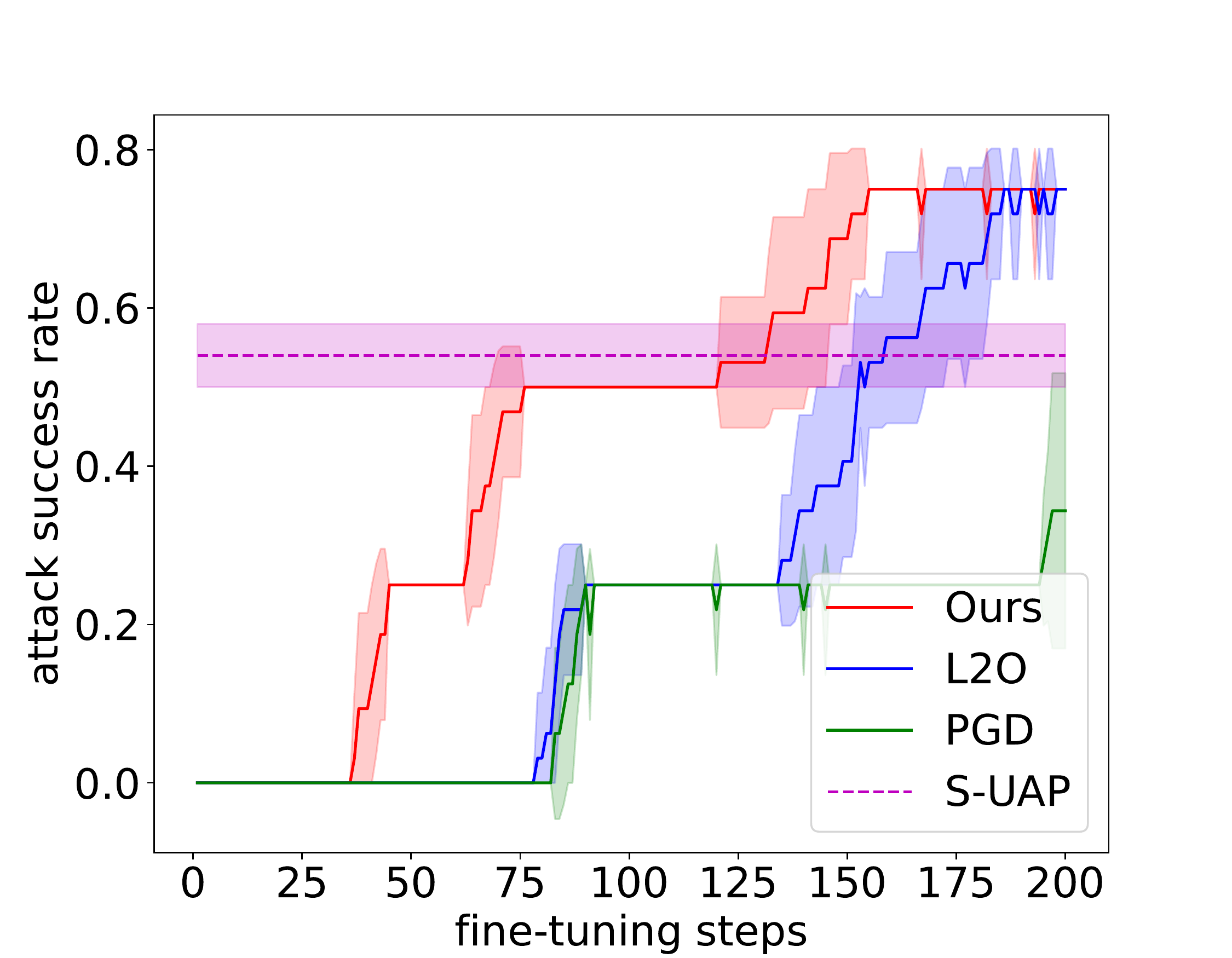}
\\ \midrule
%\makecell*[c]{ \centering trained on MNIST \\ \& CIFAR-10 }
\multicolumn{1}{c|}{\parbox[t][-1.2in][c]{0.4in}{ \centering \small{\texttt{L2}: MNIST + CIFAR-10}}} %{\footnotesize (Incongruous tasks)}}
&   %\hspace*{-0.1in}
\includegraphics[width=1.5in]{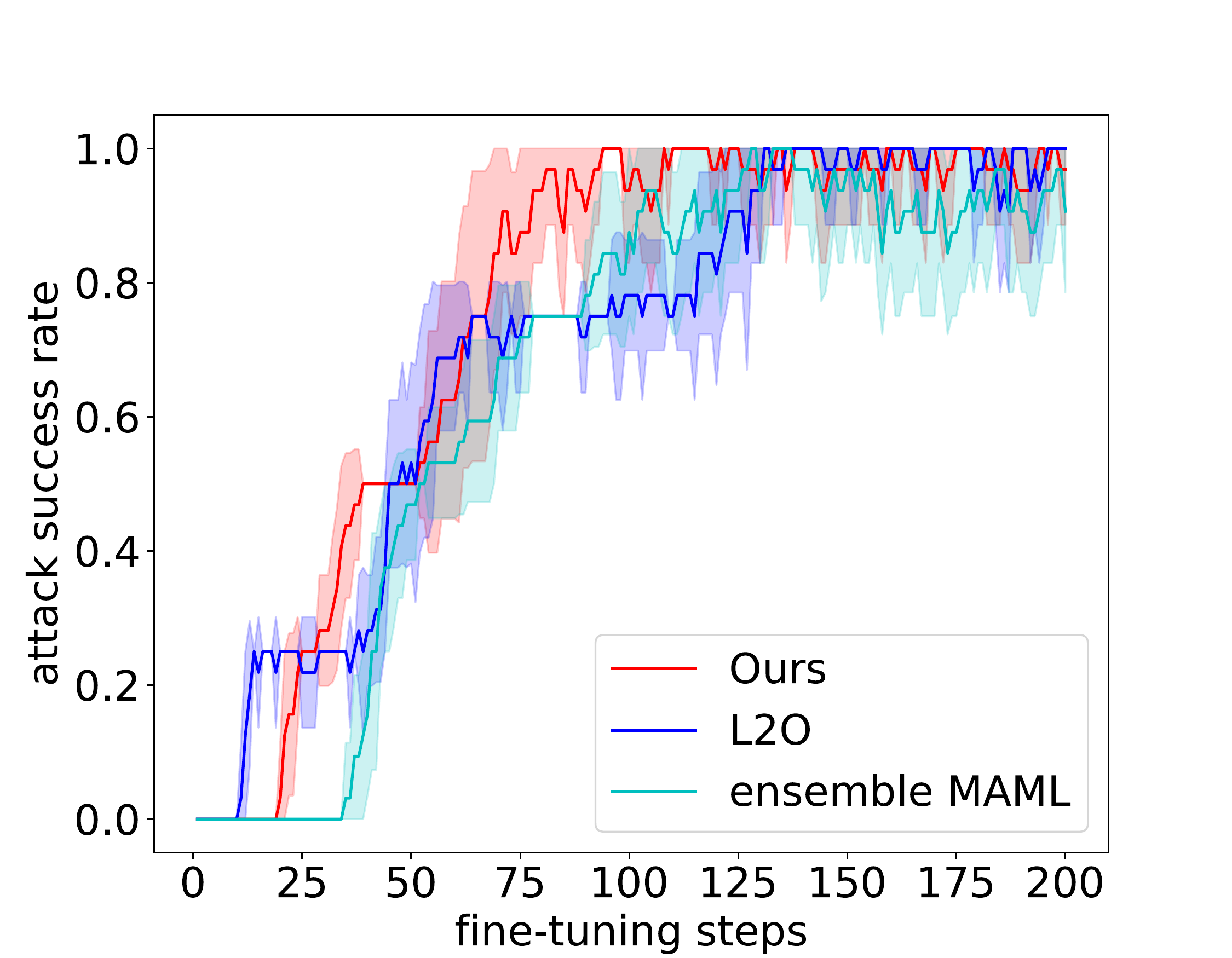} &  %\hspace*{-0.1in}
\includegraphics[width=1.5in]{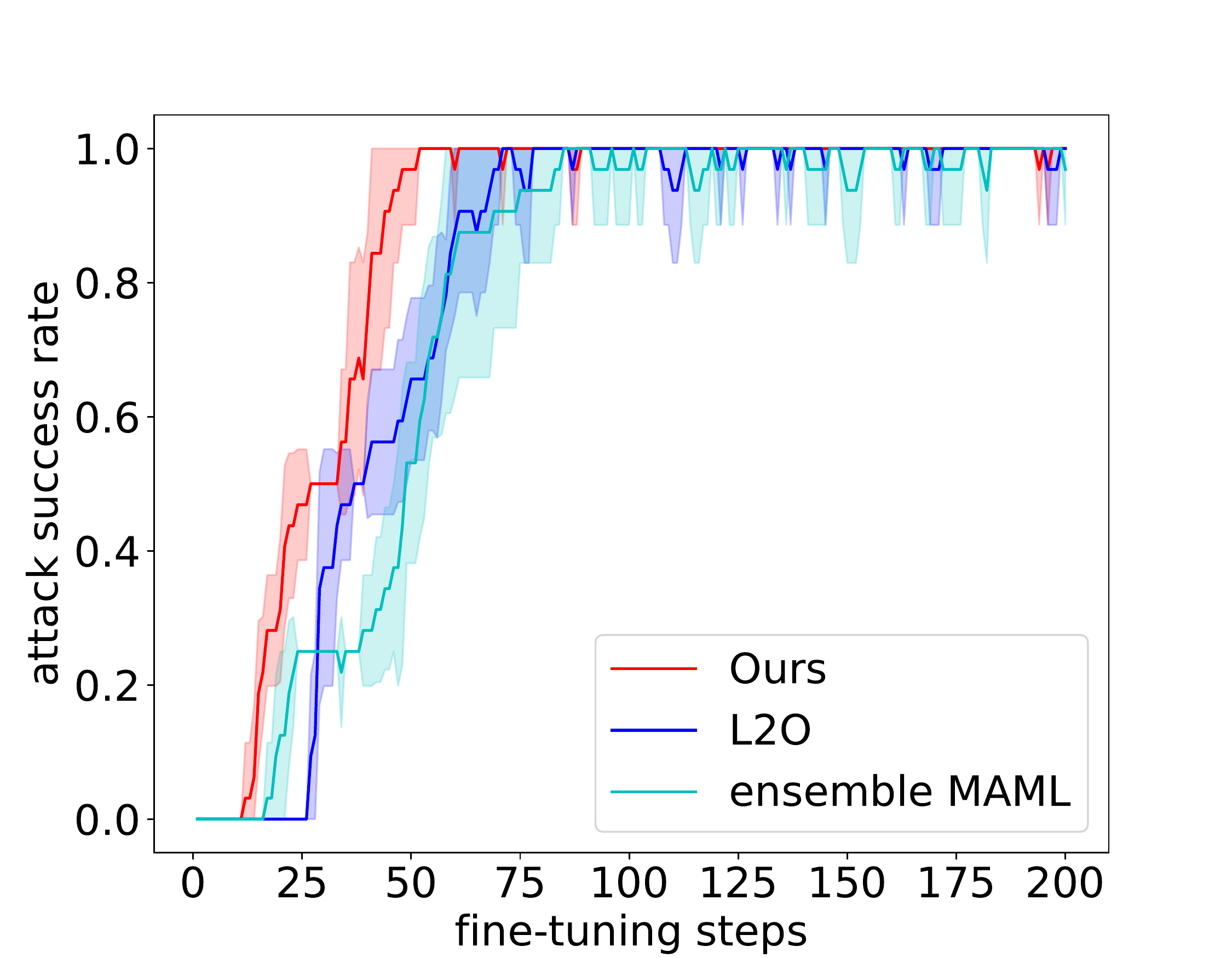} & 
\includegraphics[width=1.5in]{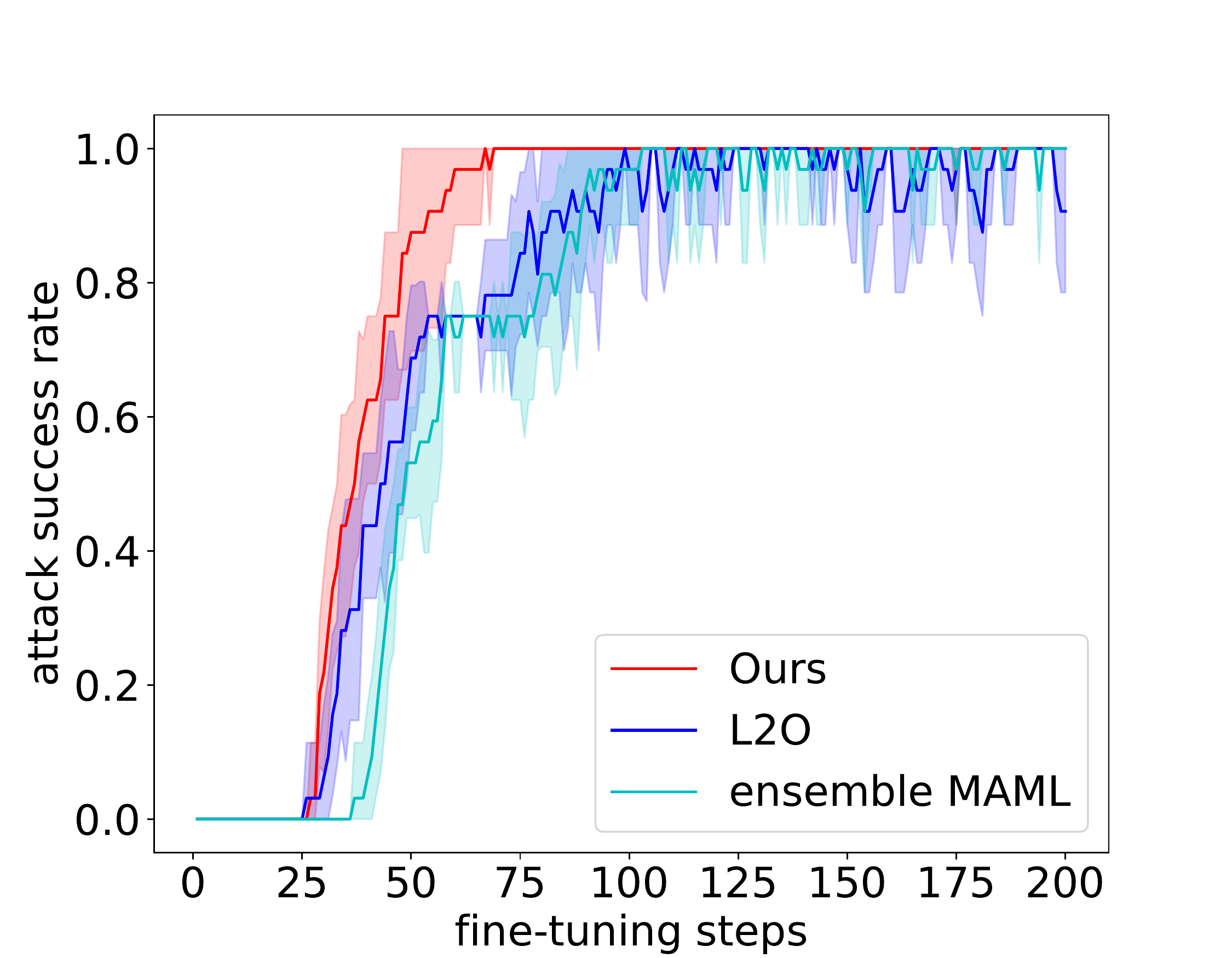} 
\\ 
\bottomrule
\end{tabular}
 \end{adjustbox}
 \vspace*{-2mm}
\caption{Aggregated ASR of UAPs generated by different schemes in various meta-learning \& meta-testing settings ({\em higher is better}). Each cell presents results for schemes meta-learnt with tasks from row-specific image source(s) and meta-tested with tasks from column-specific image source(s) with increasing UAP generation steps. S-UAP does not have an iterative process and thus we show the final ASR (dashed).
}
\label{fig:rq2}
\vspace{-10pt}
\end{figure*}

\noindent
{\em Can LFT outperform standard non-meta-learning schemes for few-shot UAP generation tasks, even for tasks from images sources \& victim models not encountered at meta-learning?}

To answer this, we meta-learn only with tasks generated from MNIST (with LeNet-5 as the victim model), and perform the meta-testing on tasks generated from (i)~MNIST with LeNet-5, (ii)~CIFAR-10 with LeNet-7, and (iii)~CIFAR-10 with VGG-11, where (ii) and (iii) involve image sources and victim models not encountered during the meta-learning. We summarize the results in \textbf{Table~\ref{tab:rq1}}. Note that MAML is only applicable in case (i). For case (i) where meta-testing tasks are from the image source used for meta-learning, all meta-learning schemes achieve 100\% ASR compared to the 63\% for S-UAP (PGD is much lower). This demonstrates the improved attack generalizability from meta-learning. For image sources not encountered during meta-learning (cases (ii) \& (iii)), LFT still continues to outperform S-UAP by 8-10\%. LFT outperforms L2O by over 10\%, highlighting the gain from the fast adaptation gradient.

\noindent
{\em Can LFT effectively meta-learn with multiple image sources and improve ASR \& convergence with lower UAP $\ell_1$ norms?}

Here, we consider meta-learning with two meta-learning setups (\texttt{L1})~only tasks from MNIST and (\texttt{L2})~meta-learning with tasks from both MNIST and CIFAR-10 (MNIST+CIFAR-10), and three meta-testing setups with tasks from (\texttt{T1})~MNIST, (\texttt{T2})~CIFAR-10, and (\texttt{T3})~both (MNIST+CIFAR-10). In  these cases, the victim architecture is LeNet-5 for MNIST and LeNet-7 (LeNet with more CONV layers) for CIFAR-10 to demonstrate the ability to work with various victim architectures. 
%the victim architecture is LeNet-5 for both image sources (leading to different models for each image source, although our proposed scheme would be applicable to other architectures as demonstrated in Table~\ref{tab:rq1}).
\textbf{Figure~\ref{fig:rq2}} presents   ASR (on unseen images $\dout_i$), aggregated over   100 UAP generation tasks, with increasing number of steps in iterative UAP generation schemes (using the seen images $\din_i$); S-UAP is a non-iterative scheme.
%; note that S-UAP is a non-iterative scheme.
Note that MAML is only applicable in the \texttt{T1} case when meta-learning with MNIST tasks in \texttt{L1} case; when meta-learning with tasks from MNIST+CIFAR-10 in \texttt{L2} case, we utilize ensemble MAML.

\textbf{Figure~\ref{fig:rq2}} indicates that our proposed LFT converges first to the best ASR, with L2O and (ensemble) MAML performing comparably to each other when applicable. As seen in Table~\ref{tab:rq1}, the non-meta-learning schemes (PGD \& S-UAP) are unable to match the ASR of the meta-learning schemes in all meta-testing cases \texttt{T1}, \texttt{T2}, \texttt{T3} -- PGD converges to less than half the ASR achieved by LFT. Note that, when meta-learning with MNIST+CIFAR-10 in case \texttt{L2}, the meta-test performance of LFT on MNIST in case \texttt{T1} is not significantly affected, indicating that meta-learning from multiple image sources {\em does not hurt} the performance of LFT. Furthermore, meta-learning from MNIST+CIFAR-10 leads to improved ASR when meta-testing with CIFAR-10 in \texttt{T2} or MNIST+CIFAR-10 in \texttt{T3}, {\em highlighting the improvement in LFT from meta-learning with multiple image sources}.

The relative performances of the meta-learning based schemes are further summarized in \textbf{Table~\ref{tab:rq2}}. These results indicate that, compared to other meta-learning baselines, LFT achieves better ASR with limited number of steps (5-18\% improved ASR), converges to 100\% ASR faster (24-37\% reduction in number of steps), and does so while producing perturbations with smaller $\ell_1$ norms (up to 10\% smaller). We present further results in Appendix~\ref{appendix: l1norm} which show that LFT produces smaller $\ell_1$ norms than PGD as well. We also study the effect of the number of tasks available per image source for meta-learning in Appendix~\ref{appendix:num-tasks}.

%\todo[inline]{\\
%   - Add note about $\ell_1$ results with PGD in appendix\\
%  - Add note about amount of tasks needed in appendix\\
%  - Point to visualizations of perturbations in appendix\\
%}
%
\paragraph{Summary of evaluation.}
We mitigate challenge \hyperlink{c1}{\texttt{C1}} by demonstrating improved attack generalizability (improved ASR on unseen images) with meta-learning. Improved convergence (to 100\% ASR) of LFT with smaller perturbation strength mitigates challenge \hyperlink{c2}{\texttt{C2}}. By demonstrating improved ASR for LFT (i) by meta-learning with tasks from different image sources, and (ii) when handling UAP generation tasks from image sources not encountered during meta-learning, we mitigate challenge \hyperlink{c3}{\texttt{C3}}.

% % %\vspace*{-0.00in}
\vspace*{-2mm}
\section{Conclusion} \label{sec:conc}
In this paper, we focus on the various challenges in UAP generation, and present a meta-learning scheme LFT that extends MAML with the learning-to-optimize framework. This LFT improves the attack universality or generalizability of UAPs generated from a small set of images, and can be meta-learned from multiple image sources and be applied to image sources and architectures not seen during meta-learning, thereby extending the universality of UAP generators.

\section{Acknowledgments}
This work is partly supported by the National Science Foundation Award CNS-1932351.

\begin{comment}
In this paper, we generalize MAML to {\em incongruous few-shot learning} with the proposed \lft.  %replacing the hand-designed optimizer in the inner fine-tuning loop of MAML with a LFT
%learned fine-tuner (LFT)
%in the form of a \rnn. 
%recurrent neural network (\rnn).
% \textit{learning to meta-learn (L2ML)}, in which a  hand-designed optimizer at the inner loop of  MAML is replaced by learning to optimize (L2O) oracle, in the form of a recurrent neural network (RNN).
We show that  {\lft} can be meta-learned across incongruous tasks and  applied to few-shot UAP generation.
%We show that L2ML is is general which can handle   incongruous learning tasks and allow for multi-step fine-tuning without the need of more than second-order derivatives during meta-training.
We also theoretically quantify the difference between our proposed meta-learning scheme and L2O, highlighting why our proposed meta-learning scheme would outperform L2O for incongruous few-shot learning.
%which sheds light on generalizing L2O to meta-learning.
Empirically, we consider a novel application of meta-learning to generate UAPs and show the superior performance of {\lft}s over the state-of-the-art meta-learners.
% We also show how our {\lft}s are outperform existing meta-learning schemes for incongruous few-shot classification and regression (when applicable).
\end{comment}


%\clearpage
%\newpage
{\small 
\bibliographystyle{named}
\bibliography{ijcai22}
}
\clearpage
\newpage

\appendix
\setcounter{section}{0}
\setcounter{figure}{0}
\setcounter{equation}{0}
\makeatletter 
\renewcommand{\thefigure}{A\@arabic\c@figure}
\makeatother
\setcounter{table}{0}
\renewcommand{\thetable}{A\arabic{table}}
\renewcommand{\theequation}{S\arabic{equation}}
\setcounter{algorithm}{0}
\renewcommand{\thealgorithm}{A\arabic{algorithm}}
\onecolumn

%\aistatstitle{Supplementary Material}
\section*{\Large{\textbf{Appendix}}
%of \textit{Learned Fine-Tuner for Incongruous Few-Shot Adversarial Learning}
}

\section{Gradients of MAML loss with respect to \rnn parameters} \label{asec:rnn-grad}
Based on $\mathbf{G}^{(k)} = \frac{ \partial \btheta^{(k)} }{ \partial \bphi } \in \mathbb R^{d \times |\bphi|}$ and \eqref{eq: RNN_update}, we obtain
\begin{align}\label{eq: Gk_update_app}
     \mathbf G^{(k)} & = \mathbf G^{(k-1)} -  \pphi{\Delta \btheta^{(k)}}.
\end{align}
For ease of presentation, we   use $ \rnn_{\bphi} (g(\btheta^{(k-1)}), \bh_{k-1})$ to represent $\Delta \btheta^{(k)} \in \mathbb R^d$, and the  RNN output $\bh^{(k)}$ of $\rnn_{\bphi}  $ is omitted when its meaning can clearly be inferred from the context. We then have
% \begin{align}
%   \pphi{\Delta \btheta^{(k)}}  = &  \pphi{ \rnn_{\bphi} (g(\btheta^{(k-1)}), \bh_{k-1}) } \label{eq: term_Delta_theta} \\
%   = &   \left(\pphi{g(\btheta^{(k-1)})}\right)^\top \left.\frac{ \partial \rnn_{\bphi}( g, h ) }{ \partial g }\right|_{g = g(\btheta^{(k-1)}), h = \bh_{k-1}}  \label{eq: term_gtheta}
%   \\
%   &  + \left(\pphi{\bh_{k-1}}\right)^\top \left.\frac{ \partial \rnn_{\bphi}( g, h ) }{ \partial h }\right|_{g = g(\btheta^{(k-1)}), h = \bh_{k-1}}  \label{eq: term_h}
%   \\
%         & + \left. \pphi{\rnn_{\bphi}(g, h)} \right|_{g = g(\btheta^{(k-1)}), h = \bh_{k-1}},  \label{eq: term_phi}
%         %\\
% % \pphi{\bh_{k-1}}  & = \left. \pphi{\bh_{k-1}(C_1, C_2)} \right|_{C_1 = g(\btheta^{(k-2)}), C_2 = \bh_{k-2}} \\
% % & \quad +
% %  \left( \pphi{g(\btheta^{(k-2)})} \right)^\top \left. \frac{\partial \bh_{k-1}(C_1, C_2) }{C_1} \right|_{C_1 = g(\btheta^{(k-2)}),  C_2 = \bh_{k-2}} \\
% % & \quad + \left(\pphi{\bh_{k-2}}\right)^\top \left. \frac{\partial \bh_{k-1}(C_1, C_2) }{C_2} \right|_{C_1 = g(\btheta^{(k-2)}),  C_2 = \bh_{k-2}}
% \end{align}
\begin{align}
  \pphi{\Delta \btheta^{(k)}}  = &  \pphi{ \rnn_{\bphi} (g(\btheta^{(k-1)}), \bh_{k-1}) } \label{eq: term_Delta_theta} \\
  = &  % \left( \pphi{g(\btheta^{(k-1)})}\right )^\top
  %\left.
  \frac{ \partial \rnn_{\bphi}( g(\btheta^{(k-1)}), \bh_{k-1} ) }{ \partial g }
  %\right|_{g = g(\btheta^{(k-1)}), h = \bh_{k-1}} 
  \cdot
  \pphi{g(\btheta^{(k-1)})}
  \label{eq: term_gtheta}
  \\
  &  + %\left(\pphi{\bh_{k-1}}\right)^\top
  %\left.
  \frac{ \partial \rnn_{\bphi}( g(\btheta^{(k-1)}), \bh_{k-1} ) }{ \partial \bh }
  %\right|_{g = g(\btheta^{(k-1)}), h = \bh_{k-1}}  
  \cdot \pphi{\bh_{k-1}} \label{eq: term_h}
  \\
        & + \left. \pphi{\rnn_{\bphi}(g, h)} \right|_{g = g(\btheta^{(k-1)}), h = \bh_{k-1}},  \label{eq: term_phi}
        %\\
% \pphi{\bh_{k-1}}  & = \left. \pphi{\bh_{k-1}(C_1, C_2)} \right|_{C_1 = g(\btheta^{(k-2)}), C_2 = \bh_{k-2}} \\
% & \quad +
%  \left( \pphi{g(\btheta^{(k-2)})} \right)^\top \left. \frac{\partial \bh_{k-1}(C_1, C_2) }{C_1} \right|_{C_1 = g(\btheta^{(k-2)}),  C_2 = \bh_{k-2}} \\
% & \quad + \left(\pphi{\bh_{k-2}}\right)^\top \left. \frac{\partial \bh_{k-1}(C_1, C_2) }{C_2} \right|_{C_1 = g(\btheta^{(k-2)}),  C_2 = \bh_{k-2}}
\end{align}
where the equality holds by chain rule \cite{petersen2008matrix},
$\cdot$ denotes a matrix product that the chain rule obeys,
and 
the term \eqref{eq: term_phi} denotes the derivative w.r.t. $\bphi$ by fixing $g(\btheta^{(k-1)})$ and $ \bh_{k-1}$ as constants.
% The term \eqref{eq: term_gtheta} can further be simplified to \begin{align} \label{eq: term_gtheta_v2_ori}
%     \frac{\partial g(\btheta^{(k-1)})}{\partial \bphi} = \left ( 
%      \frac{\partial \btheta^{(k-1)}}{\partial \bphi}
%  \right )^\top\frac{\partial g(\btheta^{(k-1)})}{\partial \btheta} = \left ( \mathbf G^{(k-1)} \right )^\top \frac{\partial g(\btheta^{(k-1)})}{\partial \btheta}. 
% \end{align}
\begin{align} \label{eq: term_gtheta_v2}
    \frac{\partial g(\btheta^{(k-1)})}{\partial \bphi} = 
    %( 
      \frac{\partial g(\btheta^{(k-1)})}{\partial \btheta} \cdot
  %)^\top
 \frac{\partial \btheta^{(k-1)}}{\partial \bphi}
= 
%\left ( 
\frac{\partial g(\btheta^{(k-1)})}{\partial \btheta} \cdot
%\right )^\top 
\mathbf G^{(k-1)}.
\end{align}

Next, we simplify the term \eqref{eq: term_h}. Let 
$\mathbf H^{(k)} =\pphi{\bh_{k}} \in \mathbb R^{|\bh_k| \times |\bphi|}$. Note that $\bh_k$ depends on $\bphi, g(\btheta^{(k-1)}), \bh_{k-1}$. So we write $\bh_k = \Pi(\bphi, g(\btheta^{(k-1)}), \bh_{k-1})$

\begin{align}
    \mathbf H^{(k)} = \pphi{\bh_{k}} = &
    \pphi{}\Pi(\bphi, g(\btheta^{(k-1)}), \bh_{k-1}) \nonumber \\
%    = &\left. \pphi{\bh_{k}(C_1, C_2)} \right|_{C_1 = g(\btheta^{(k-1)}), C_2 = \bh_{k-1}} \nonumber  \\
    = &\left. \pphi{\Pi(\bphi, g, h)} \right|_{g = g(\btheta^{(k-1)}), h = \bh_{k-1}} \nonumber  \\
%&  +
% \left( \pphi{g(\btheta^{(k-1)})} \right)^\top \left. \frac{\partial \bh_{k}(C_1, C_2) }{C_1} \right|_{C_1 = g(\btheta^{(k-1)}),  C_2 = \bh_{k-1}} \nonumber  \\
%&  + \left(\pphi{\bh_{k-1}}\right)^\top \left. \frac{\partial \bh_{k}(C_1, C_2) }{C_2} \right|_{C_1 = g(\btheta^{(k-1)}),  C_2 = \bh_{k-1}} \nonumber \\
%\overset{\eqref{eq: term_gtheta_v2}}{=} & \left. \pphi{\bh_{k}(C_1, C_2)} \right|_{C_1 = g(\btheta^{(k-1)}), C_2 = \bh_{k-1}} \nonumber \\
%& + \left ( \frac{\partial g(\btheta^{(k-1)})}{\partial \btheta}   \right )^\top \mathbf G^{(k-1)} \left. \frac{\partial \bh_{k}(C_1, C_2) }{C_1} \right|_{C_1 = g(\btheta^{(k-1)}),  C_2 = \bh_{k-1}} \nonumber \\
%& + \left ( \mathbf H^{(k-1)} \right )^{\top} \left. \frac{\partial \bh_{k}(C_1, C_2) }{C_2} \right|_{C_1 = g(\btheta^{(k-1)}),  C_2 = \bh_{k-1}}. \label{eq: term_h_v2}
&  +
 %\left( \pphi{g(\btheta^{(k-1)})} \right)^\top
% \left. 
 \frac{\partial \Pi(\bphi, g(\btheta^{(k-1)}) , \bh_{k-1})}{\partial g} 
 %\right|_{g = g(\btheta^{(k-1)}), h = \bh_{k-1}}
 \cdot \pphi{g(\btheta^{(k-1)})}
 \nonumber  \\
&  + %\left(
%\right)^\top 
%\left. 
\frac{\partial  \Pi(\bphi,  g(\btheta^{(k-1)}),  \bh_{k-1}) }{\partial \bh} 
%\right|_{g = g(\btheta^{(k-1)}), h = \bh_{k-1}}
\cdot \pphi{\bh_{k-1}}
\nonumber \\
\overset{\eqref{eq: term_gtheta_v2}}{=} & \left. \pphi{\Pi(\bphi, g, h)} \right|_{g = g(\btheta^{(k-1)}), h = \bh_{k-1}} \nonumber \\
& + %\left ( \frac{\partial g(\btheta^{(k-1)})}{\partial \btheta}   \right )^\top \mathbf G^{(k-1)}
 %\left. 
 \frac{\partial \Pi(\bphi, g(\btheta^{(k-1)}), \bh_{k-1} ) }{\partial g} 
 %\right|_{g = g(\btheta^{(k-1)}),  h = \bh_{k-1}} 
\cdot 
 \frac{\partial g(\btheta^{(k-1)})}{\partial \btheta}      \mathbf G^{(k-1)}
\nonumber \\
& + 
%\left. 
\frac{\partial \Pi(\bphi,  g(\btheta^{(k-1)}), \bh_{k-1}) }{\partial \bh}
%\right|_{g = g(\btheta^{(k-1)}), h = \bh_{k-1}} 
\cdot  \mathbf H^{(k-1)}  . \label{eq: term_h_v2}
\end{align}
%meaning is 

Substituting \eqref{eq: term_gtheta_v2} and \eqref{eq: term_h_v2} into \eqref{eq: term_Delta_theta}, we can then express \eqref{eq: Gk_update_app} as:
\begin{align}
    \mathbf G^{(k)}  = & \mathbf G^{(k-1)} -   
    %\left.
    \frac{ \partial \rnn_{\bphi}(  g(\btheta^{(k-1)}) , \bh_{k-1} ) }{ \partial g }
    %\right|_{g = g(\btheta^{(k-1)}), h = \bh_{k-1}}
    \cdot  \frac{\partial g(\btheta^{(k-1)})}{\partial \btheta}   \cdot \mathbf G^{(k-1)}
    \nonumber \\
    &  {-}  %\left.
    \frac{ \partial \rnn_{\bphi}( g(\btheta^{(k-1)}), \bh_{k-1} ) }{ \partial \bh }
 %   \right|_{g = g(\btheta^{(k-1)}), h = \bh_{k-1}}
    \cdot  \mathbf H^{(k-1)} \nonumber \\
    & {-} \left. \pphi{\rnn_{\bphi}(g, h)} \right|_{g = g(\btheta^{(k-1)}), h = \bh_{k-1}}, \label{eq: Gk_v2}
\end{align}
where $\mathbf H^{(k-1)}$ is determined by the recursion   \eqref{eq: term_h_v2}. 
% We note that there were typos  in the submitted version of (7) (in the main paper), where the last two terms of   \eqref{eq: Gk_v2} were missing. This could lead to confusion. In the supplementary material, we have corrected these typos in \eqref{eq: Gk_v2}, and have updated the description of the main paper (lines 175-180) for clarity.

%We note that there was a typo in the submitted version of \eqref{eq: Gk}, where the last two terms of   \eqref{eq: Gk_v2} were missing.
It is clear from \eqref{eq: term_h_v2} and \eqref{eq: Gk_v2} that  the second order derivative would at most be involved due to the presence of  $\frac{\partial g(\btheta^{(k-1)})}{\partial \btheta}$ if $g(\btheta^{(k-1)})$ is specified by the first-order derivative w.r.t. $\btheta$. By contrast, if it is specified by the ZO gradient estimate, then there will only be first-order derivatives involved in \eqref{eq: term_h_v2} and \eqref{eq: Gk_v2}. Lastly, we remark that the recursive forms of \eqref{eq: term_h_v2} and \eqref{eq: Gk_v2} facilitate our computation, and   $\mathbf G^{(0)} = \mathbf 0$ and $\mathbf H^{(0)} = 0$.

\section{Proof of Proposition 1} \label{asec:grad-diff-theory}

Before showing the theoretical results, we first give the following a blanket of assumptions.
\subsection{Assumptions}
%A1. Gradient Lipschitz continuity of function $f_i,\forall i$ with constant $L,\forall i,k$:
%\begin{equation}
%     \|\nabla f^{(k)}_i(\btheta)-\nabla f^{(k)}_i(\btheta')\|\le L\|\btheta^{(k)}-\btheta'^{(k)}\|
%\end{equation}

In practice, the size of data and variables are limited and the function is also bounded. To proceed, we have the following standard assumptions for quantifying the gradient difference between L2O and LFT. 

A1. We assume that gradient estimate is unbiased, i.e.,
\begin{equation}
    \left[\frac{1}{N}\sum_{i=1}^N\mathbb{E}_{\mathcal{D}^{\cdot}_i} \nabla_{\btheta_i} f_i(\btheta^{(k)}_i;\mathcal{D}^{\cdot})\right]=\frac{1}{N}\sum_{i=1}^N\grad_{\btheta_i} f_i(\btheta^{(k)}_i):=\nabla_{\btheta} f(\btheta^{(k)}_1,\ldots,\btheta^{(k)}_N),\forall k,
\end{equation}
where $\mathcal{D}^{\cdot}_i$ denotes the training/validation data sample of the $i$th task, and $\btheta$ stands for $\btheta_1,\ldots,\btheta_N$.
%\todo[inline, author=PR]{
%@Songtao: S9 looks weird -- the expectation is w.r.t. $\din_i$ but the sum inside the expectation is a sum over $i$. Also, what is $f$ and $\btheta_k$ on the RHS? Which task does it correspond to? Do we want something like $\E_{\din_i \sim \task_i} \grad f_i(\btheta_i^{(k)}; \din_i) = \grad f_i(\btheta_i^{(k)})$? Or are we trying an expectation over all tasks and not just over $\din_i \sim \task_i$?
%
%STL: the key here is that we need to remove the task related parameter since L2O is a pure learning process without saying anything about task.
%
%PR: Since all the $\grad$ are with respect to $\bphi$, can we have the final defined variables as $\grad_{\bphi}^{\mathtt{tr}}$. Because we have not defined what this universal $f$ \& $\btheta$ is.
%
%STL: Sure. }

A2. We assume that the gradient estimate has bounded variance for both $\din_i, \dout_i,\forall i,k$, i.e.,
\begin{equation}
    \mathbb{E}_{\mathcal{D}^{(\cdot)}_i} \left[\left\|\nabla_{\btheta_i} f_i(\btheta^{(k)}_i;\mathcal{D}^{\cdot}_i)-\nabla_{\btheta} f(\btheta^{(k)}_1,\ldots,\btheta^{(k)}_N)\right\|^2\right]\le\sigma^2,\forall i,k.
\end{equation}

The same assumption is also applied for $\partial \btheta^{(k)}_i/\partial \bphi$:
\begin{align}
    & \mathbb{E} \left[\frac{1}{N}\sum^N_{i=1}\frac{\partial  (\btheta^{(k)}_i; \mathcal{D}^{\cdot}_i)}{\partial\bphi}\right]=\frac{1}{N}\sum_{i=1}^N\frac{\partial \btheta^{(k)}_i}{\partial \bphi}:=  \frac{\partial f(\btheta^{(k)}_1,\ldots,\btheta^{(k)}_N)}{\partial \bphi},\forall i,k,
    \\
    & \mathbb{E} \left[\left\|\frac{\partial  (\btheta^{(k)}_i;\mathcal{D}^{\cdot}_i)}{\partial\bphi}-\frac{\partial f(\btheta^{(k)}_1,\ldots,\btheta^{(k)}_N)}{\partial \bphi}\right\|^2\right]\le\sigma^2,\forall i,k.
\end{align}

A3. We assume that the size of gradient is uniformly upper bounded, i.e., $\mathbb{E}\|\nabla f_i(\btheta^{(k)}_i,\mathcal{D}^{(\cdot)}_i)\|\le G$, $ \mathbb{E}\|\frac{\partial (\btheta^{(k)}_i,\mathcal{D}_i^{(\cdot)})}{\partial \bphi}\|\le G,\forall i,k$.

%Define
%\begin{equation}
%     F(f_{\task}; \dout_{\task})=\sum_{i=1}^N \E_{\dout_{\task_i}\sim\task_i} F( f_{\task_i}; \dout_{\task_i}),
%\end{equation}

%\subsubsection{Bounded Variance}
%\begin{equation}
%    \mathbb{E}\|\nabla_{\bphi}  F(f_{\task}; \dout_{\task})-\nabla_{\bphi}F(f_{\task_i}; \dout_{\task_i})\|^2\le\sigma^2
%\end{equation}
%\begin{equation}
%    \mathbb{E}\|\nabla_{\bphi}  F(f_{\task_i})-\nabla_{\bphi}F(f_{\task_i}; \dout_{\task_i})\|^2\le\sigma'^2
%\end{equation}
\emph{Proof}.
Assume that A1--A3 hold. Let $w_k=\frac{1}{K}$. From the definitions of $F(\bphi)$ and $\widehat{F}(\bphi)$, we have
\begin{align}
\notag
&\|\nabla_{\bphi} \widehat{F}(\bphi)-\nabla_{\bphi} F(\bphi)\|^2
\\
= &\left\|\frac{1}{N}\sum^N_{i=1}\frac{1}{K}\sum_{k=1}^K  \E_{\din_i \dout_i} \nabla_{\bphi} f( \btheta_i^{(k)};\dout_i)- \nabla_{\bphi} f(\btheta^{(k)}_1,\ldots,\btheta^{(k)}_N)\right\|^2
\\
\mathop{\le}\limits^{(a)} &\frac{1}{K}\sum^K_{k=1}\left\|\E_{\din_i \dout_i}\frac{1}{N}\sum^N_{i=1}\nabla_{\bphi} f( \btheta_i^{(k)};\dout_i)- \nabla_{\bphi} f(\btheta^{(k)}_1,\ldots,\btheta^{(k)}_N)\right\|^2
\\
\mathop{\le}\limits^{(b)} &\frac{1}{K}\sum^K_{k=1}\left\|\frac{1}{N}\sum^N_{i=1}\E_{\din_i \dout_i}\nabla_{\bphi} f( \btheta_i^{(k)}; \dout_i)- \nabla_{\bphi}  f(\btheta^{(k)}_1,\ldots,\btheta^{(k)}_N)\right\|^2
\\
\mathop{\le}\limits^{(c)} & \frac{1}{K}\sum^K_{k=1}\left\|\frac{1}{N}\sum^N_{i=1}\E_{\din_i \dout_i}\frac{\partial f( \btheta_i^{(k)};\dout_i)}{\partial \btheta^{(k)}_i}\frac{\partial (\btheta^{(k)}_i;\din_i)}{\partial \bphi}- \nabla_{\bphi}  f(\btheta^{(k)}_1,\ldots,\btheta^{(k)}_N)\right\|^2
\\\notag
\le &\frac{1}{K}\sum^K_{k=1}\frac{1}{N}\sum^N_{i=1}\Bigg\|\E_{\din_i \dout_i}\frac{\partial f( \btheta_i^{(k)}; \dout_i)}{\partial \btheta^{(k)}_i}\frac{\partial (\btheta_i^{(k)};\din_i)}{\partial \bphi}
-\frac{\partial f( \btheta^{(k)}_1,\dots, \btheta^{(k)}_N)}{\partial \btheta^{(k)}_i}\frac{\partial (\btheta^{(k)}_i;\din_i)}{\partial \bphi}
\\
&\quad+\frac{\partial f( \btheta^{(k)}_1,\dots, \btheta^{(k)}_N)}{\partial \btheta^{(k)}_i}\frac{\partial (\btheta^{(k)}_i;\din_i)}{\partial \bphi}- \nabla_{\bphi}  f(\btheta^{(k)}_1,\ldots,\btheta^{(k)}_N)\Bigg\|^2
\\
\mathop{\le}\limits^{(d)} & \frac{2G^2}{K}\sum^K_{k=1}\left(\frac{\sigma^2}{D_{\texttt{val}}}+\frac{\sigma^2}{D_{\texttt{tr}}}\right)
\\
\le &2G^2\sigma^2\left(\frac{1}{D_{\texttt{val}}}+\frac{1}{D_{\texttt{tr}}}\right)
%\\
%\mathop{\le}\limits^{(f)} &  2G^2\sigma^2\left(\frac{1}{D_{out}}+\frac{1}{D_{in}}\right)
\end{align}
where in $(a)$ we use Jensen's inequality, in $(b)$ we apply the triangle inequality, in $(c)$ we use the chain rule, $(d)$ is true because
\begin{align}
\notag
  &\left\|\E_{\din_i \dout_i}\frac{\partial f( \btheta_i^{(k)};\dout_i)}{\partial \btheta^{(k)}_i}\frac{\partial (\btheta_i^{(k)},\din_i)}{\partial \bphi}-\frac{\partial f( \btheta^{(k)}_1,\ldots, \btheta^{(k)}_N)}{\partial \btheta^{(k)}_i}\frac{\partial (\btheta_i^{(k)},\din_i)}{\partial \bphi}\right\|^2
  \\
 % =& \left\|\E_{\din_i}\E_{\dout_i|\din_i}\frac{\partial (\btheta_i,\din_i)}{\partial \bphi}\left(\frac{\partial f( \btheta_i^{(k)}; \dout_i)}{\partial \btheta}- \frac{\partial f( \btheta_i^{(k)})}{\partial \btheta}\right)\right\|^2
 % \\
  \mathop{\le}\limits^{(i)} & \left\|\E_{\dout_i}\E_{\din_i|\dout_i}\frac{\partial (\btheta_i^{(k)},\din_i)}{\partial \bphi}\right\|^2\left\|\E_{\din_i}\E_{\dout_i|\din_i}\frac{\partial f( \btheta_i^{(k)}; \dout_i)}{\partial \btheta^{(k)}_i}- \frac{\partial f( \btheta^{(k)}_1,\ldots, \btheta^{(k)}_N)}{\partial \btheta^{(k)}_i}\right\|^2
  \\
  \mathop{\le}\limits^{(ii)} & G^2\E_{\din_i}\E_{\dout_i|\din_i}\left\|\frac{\partial f( \btheta_i^{(k)}; \dout_i)}{\partial \btheta^{(k)}_i}- \frac{\partial f( \btheta^{(k)}_1,\ldots, \btheta^{(k)}_N)}{\partial \btheta^{(k)}_i}\right\|^2
  \\
  \le &\frac{G^2\sigma^2}{D_{\texttt{val}}}
\end{align}
where in $(i)$ we use Cauchy-Schwarz inequality, in $(ii)$ we use Jensen's inequality; and similarly we have
\begin{align}
\notag
 &\left\|\E_{\din_i \dout_i}\frac{\partial f( \btheta^{(k)}_1,\ldots,\btheta^{(k)}_N)}{\partial \btheta^{(k)}_i}\frac{\partial (\btheta_i^{(k)},\din_i)}{\partial \bphi}-\nabla_{\bphi} f( \btheta^{(k)}_1,\ldots, \btheta^{(k)}_N)\right\|^2
  \\
  \le& \left\|\E_{\din_i}\frac{\partial f( \btheta^{(k)}_1,\ldots,\btheta^{(k)}_N)}{\partial \btheta^{(k)}_i}\left(\frac{\partial (\btheta_i^{(k)}; \din_i)}{\partial \bphi}- \frac{\partial f(  \btheta^{(k)}_1,\ldots,\btheta^{(k)}_N)}{\partial \bphi}\right)\right\|^2
  \\
  \le & \left\|\frac{\partial f( \btheta^{(k)}_1,\ldots,\btheta^{(k)}_N)}{\partial \btheta^{(k)}_i}\right\|^2\left\|\E_{\din_i}\frac{\partial ( \btheta_i^{(k)}; \din_i)}{\partial \bphi}- \frac{\partial  f(\btheta^{(k)}_1,\ldots,\btheta^{(k)}_N)}{\partial \bphi}\right\|^2
  \\
  \le & G^2\E_{\din_i}\left\|\frac{\partial ( \btheta_i^{(k)};\din_i)}{\partial \bphi}- \frac{\partial f(  \btheta^{(k)}_1,\ldots,\btheta^{(k)}_N)}{\partial \bphi}\right\|^2
  \\
  \le &\frac{G^2\sigma^2}{D_{\texttt{tr}}}.
\end{align}

Then, we can have
\begin{align}
    &\|\nabla_{\bphi} F(\bphi)-\nabla_{\bphi} \widehat{F}(\bphi)\|
    \le \sqrt{2}G\sigma\sqrt{\frac{1}{D_\texttt{tr}}+\frac{1}{D_{\texttt{val}}}}
        \sim\mathcal{O}\left(\underbrace{\sqrt{\frac{1}{D_\texttt{tr}}+\frac{1}{D_{\texttt{val}}}}}_{\epsilon(D_{\texttt{tr}},D_{\texttt{val}})}\right).
\end{align}

Therefore, it is shown that when $D_{\texttt{tr}}$ and $D_{\texttt{val}}$ are both large, then the difference between $\nabla_{\bphi} F(\bphi)$ and $\nabla_{\bphi}\widehat{F}(\bphi)$ are quite small. If we assume that L2O converges to the stationary points in the sense the L2O algorithm is able to find a good solution for equation (9) in the main paper by optimizing $\bphi$, then it is implied that our approach will also converge to the stationary point up some error, i.e., $\epsilon(D_{\texttt{tr}},D_{\texttt{val}})$ which is small. 
$\hfill\blacksquare$

\section{$\ell_1$ norm and loss comparison} \label{appendix: l1norm}

\begin{figure*}[htb]
 \centering
   \begin{adjustbox}{max width=0.75\textwidth }
\begin{tabular}{m{1.0in}p{1.4in}p{1.4in}p{1.4in}}
\toprule
\diagbox[width=10em,trim=l]{{Meta-learning}}{{Meta-test}}
& \makecell{\centering  {\footnotesize \texttt{T1}: MNIST}}  
& \makecell{\centering  {\footnotesize \texttt{T2}: CIFAR-10}} 
& \makecell{\centering  {\footnotesize \texttt{T3}: MNIST + CIFAR-10}}
\\
\midrule
\parbox[t][-1.2in][c]{0.4in}{ \centering \small{\texttt{L1}: MNIST} }%{\footnotesize (Congruous tasks)} }
 &   %\hspace*{-0.1in}
\includegraphics[width=1.5in]{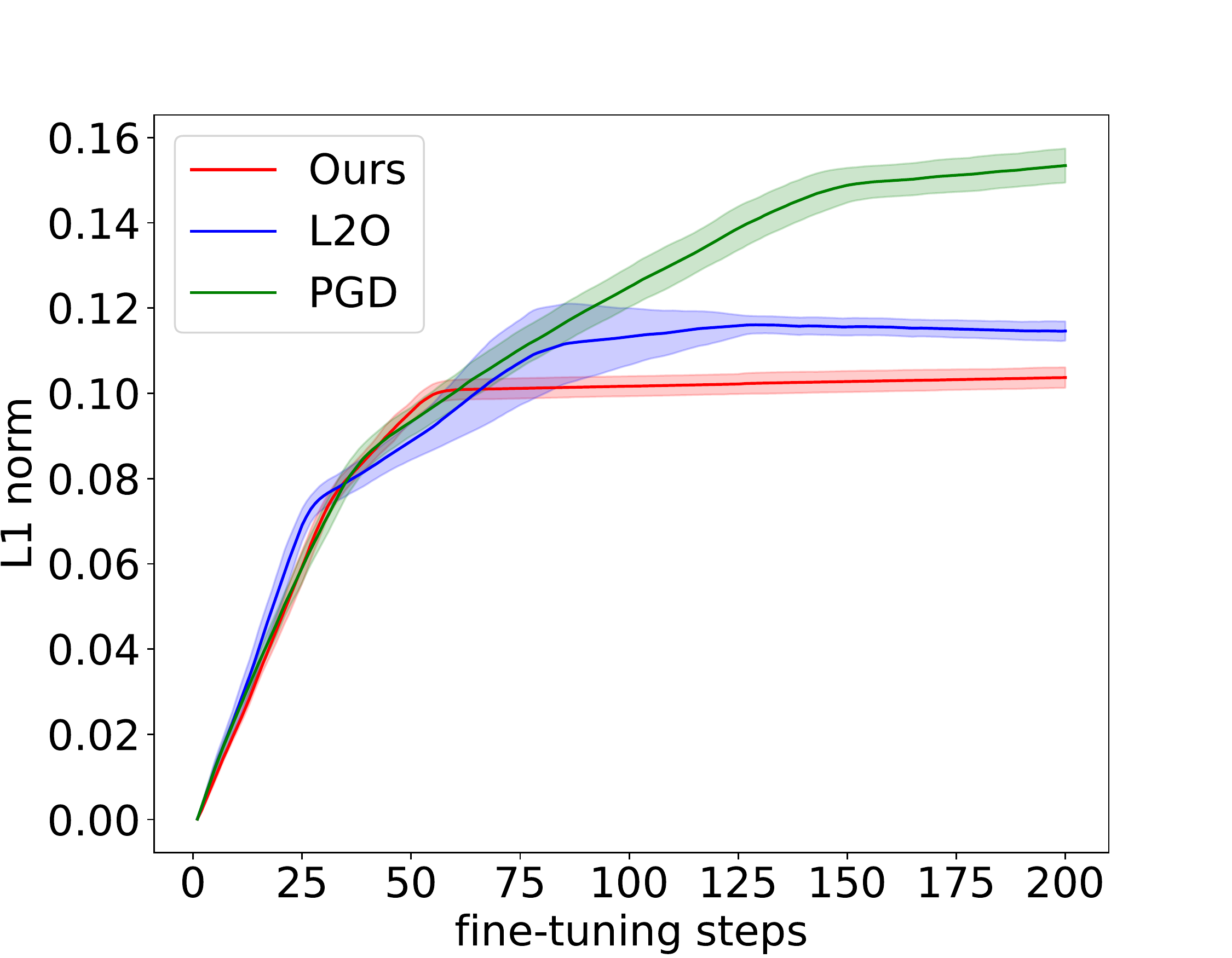}& %\hspace*{-0.1in}
\includegraphics[width=1.5in]{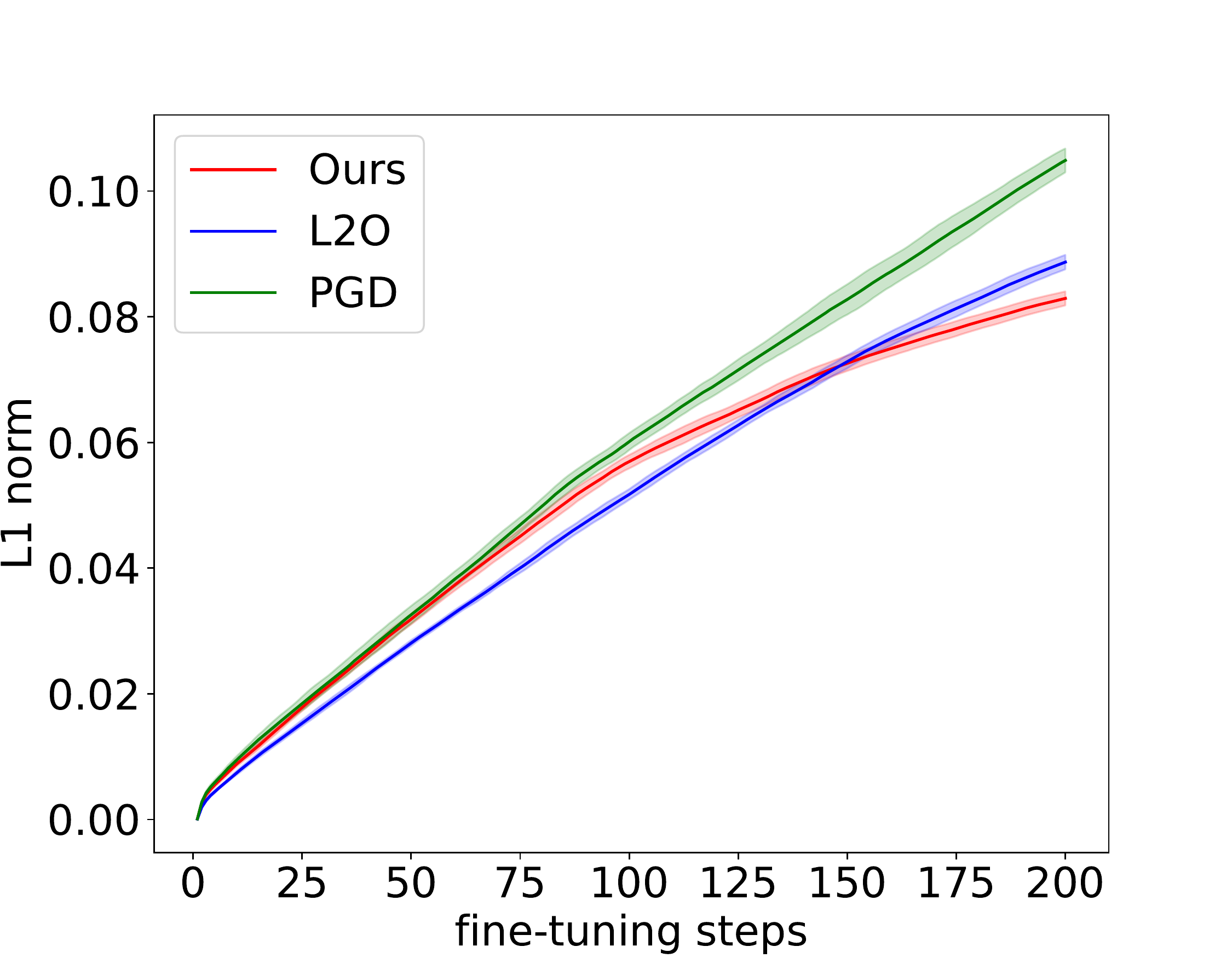}&
\includegraphics[width=1.5in]{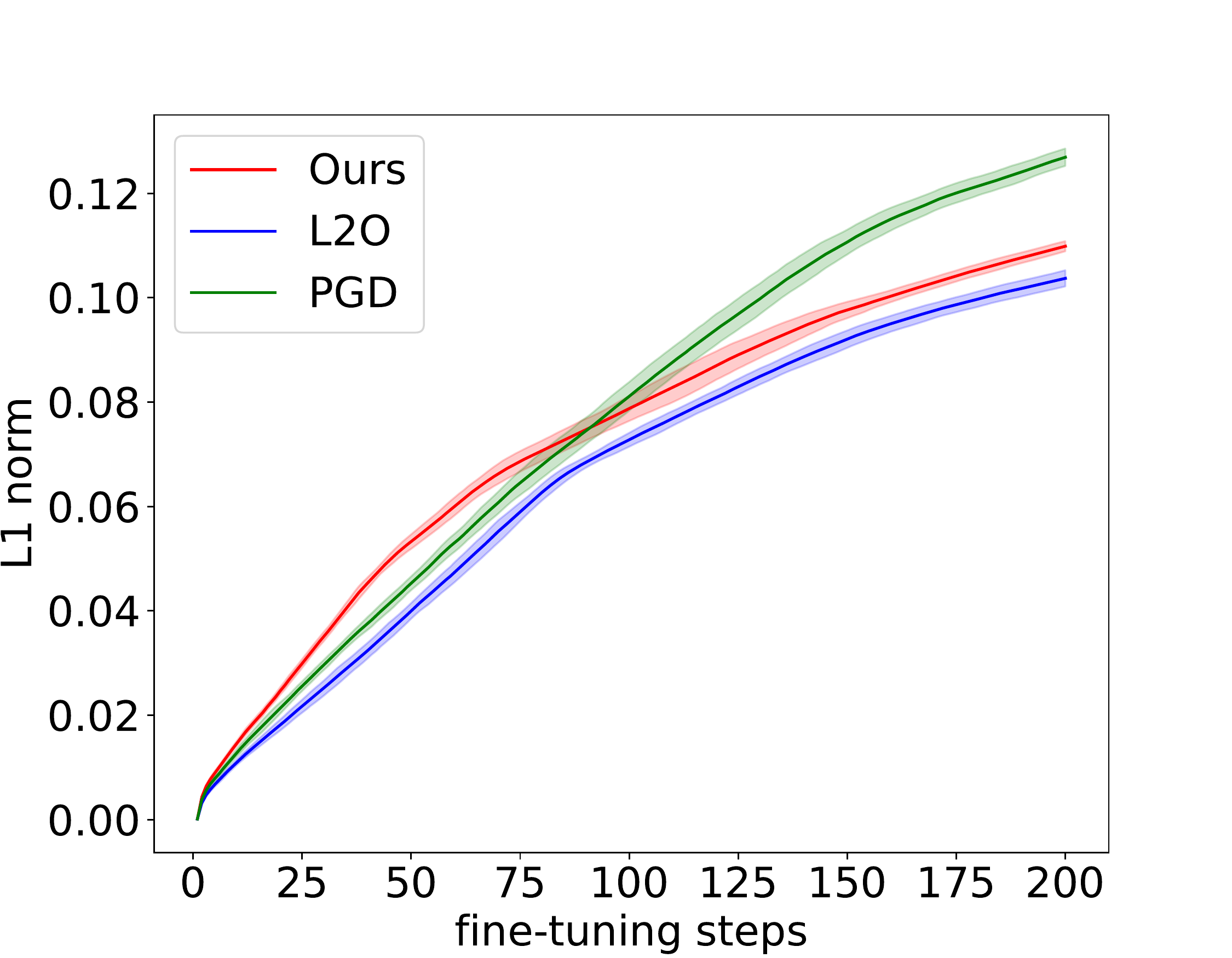}
\\ \midrule
%\makecell*[c]{ \centering trained on MNIST \\ \& CIFAR-10 }
\parbox[t][-1.2in][c]{0.4in}{ \centering  \small{\texttt{L2}: MNIST + CIFAR-10}} %{\footnotesize (Incongruous tasks)}}
&   %\hspace*{-0.1in}
\includegraphics[width=1.5in]{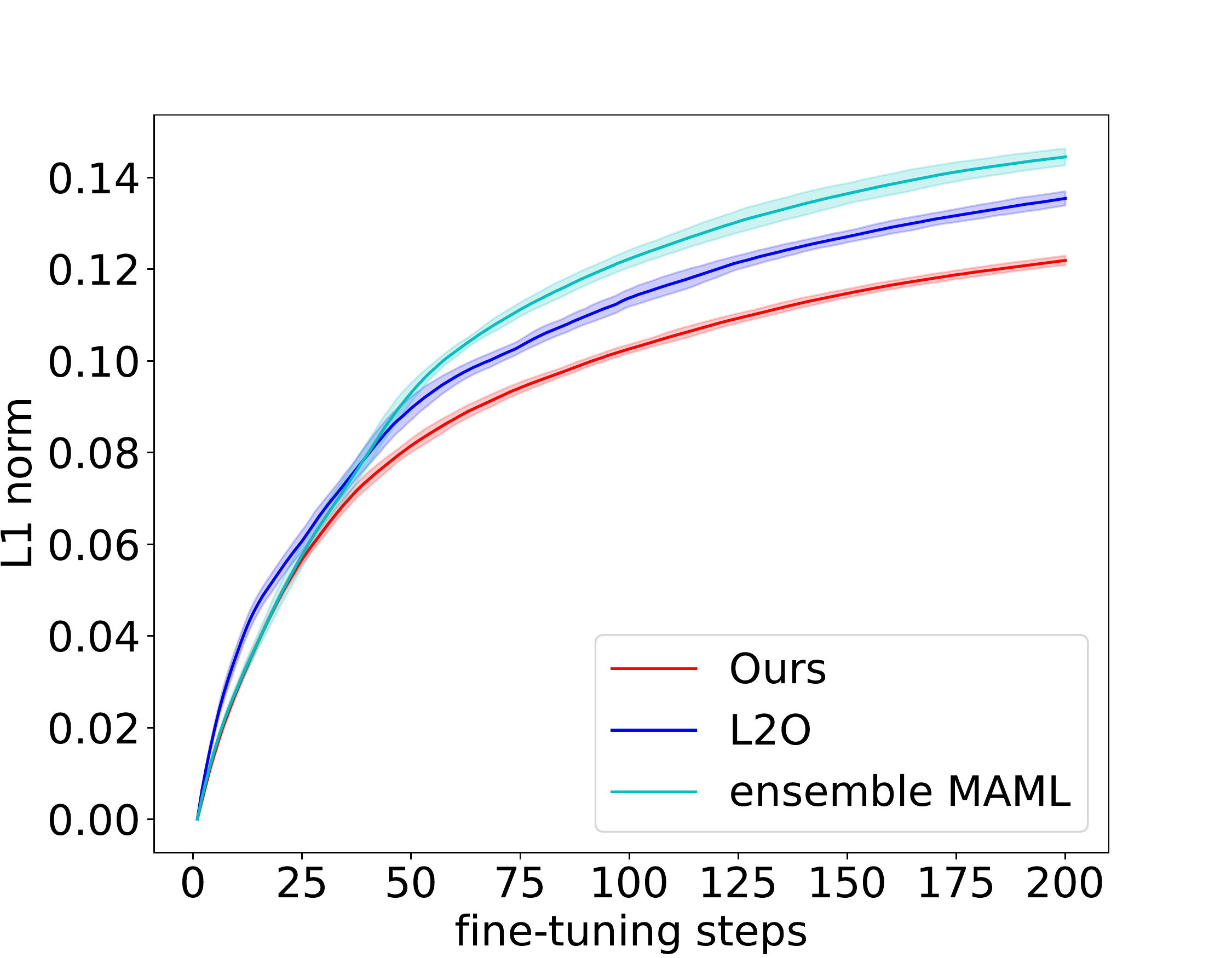} &  %\hspace*{-0.1in}
\includegraphics[width=1.5in]{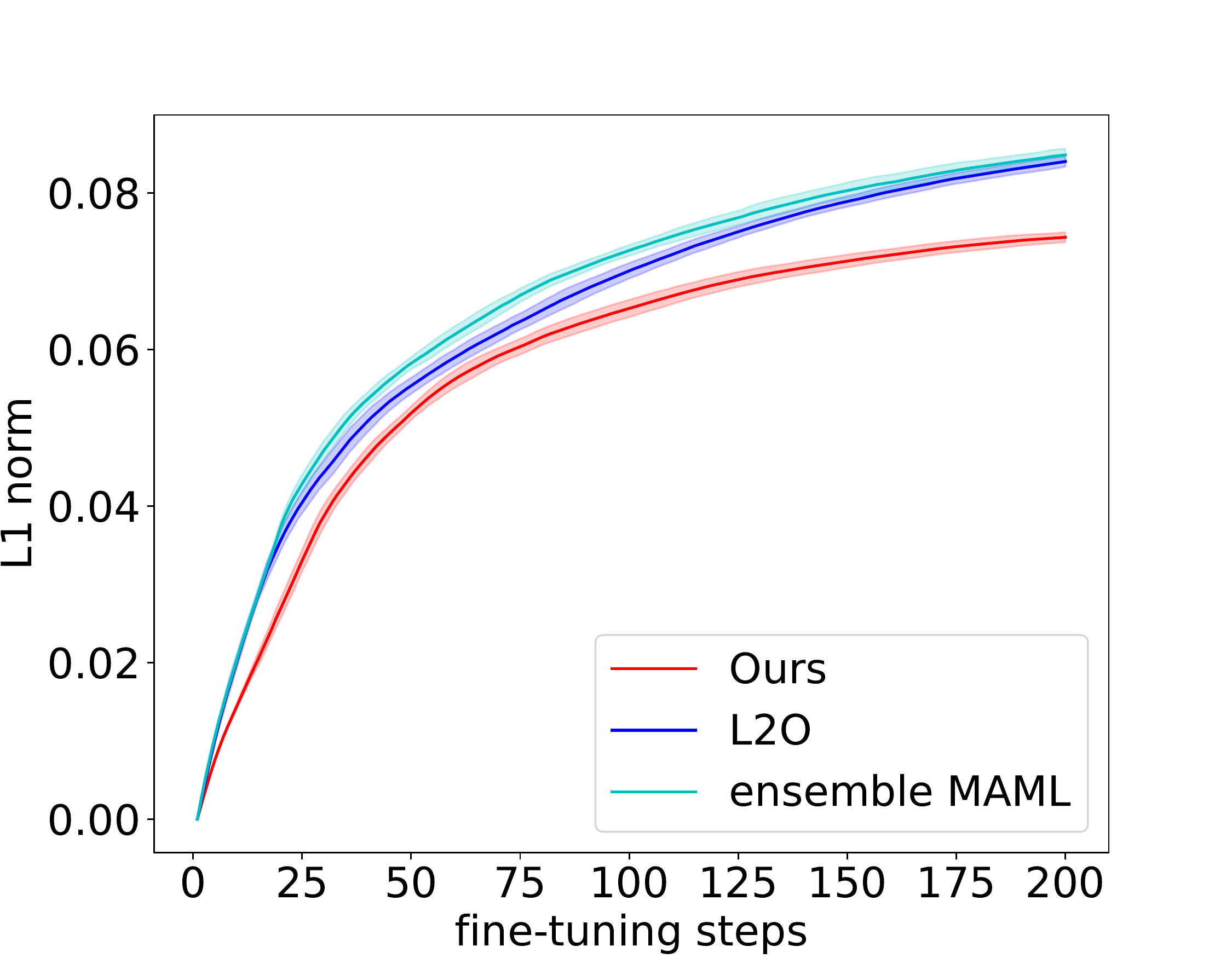} &
\includegraphics[width=1.5in]{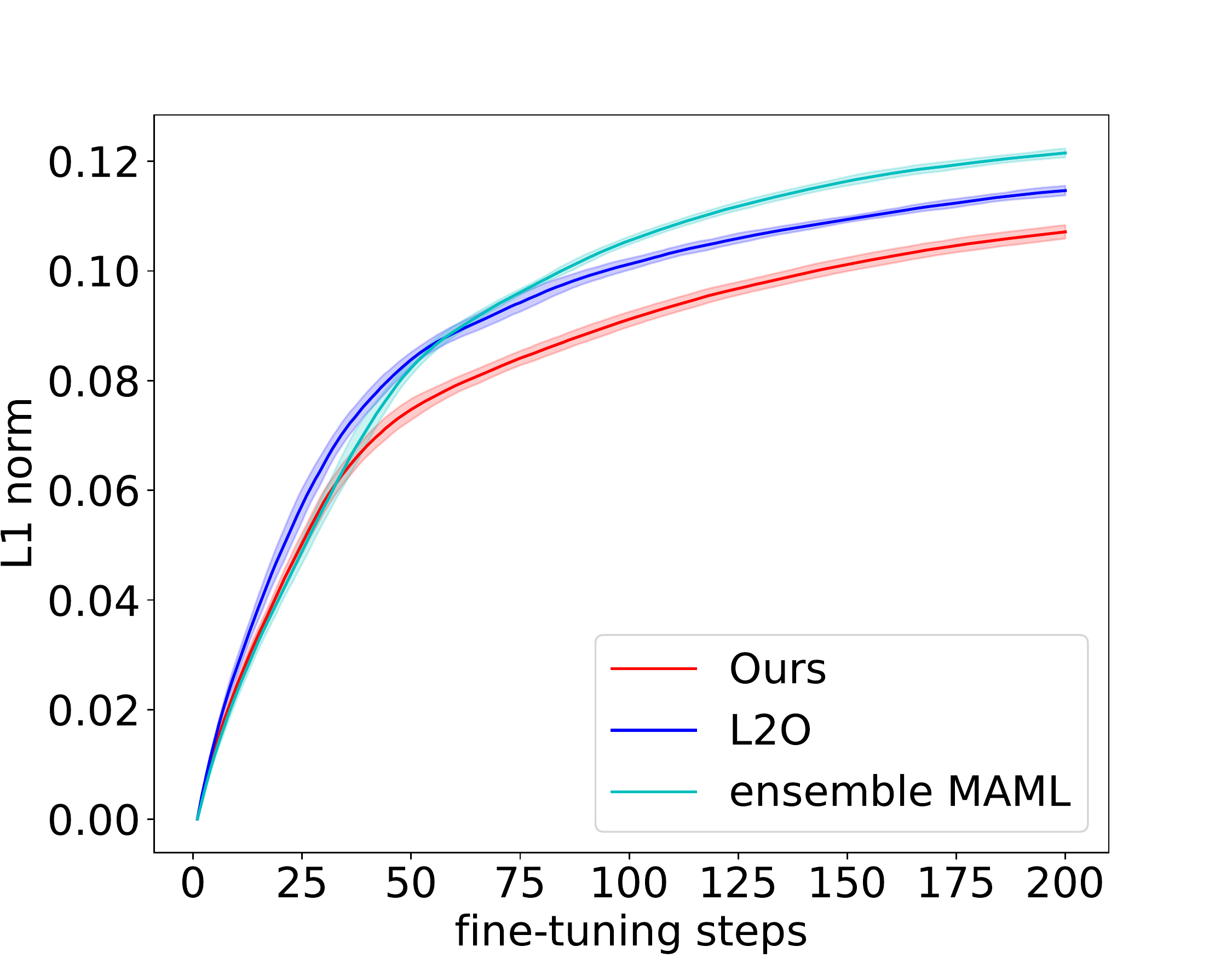}
\\
\bottomrule
\end{tabular}
 \end{adjustbox}
\caption{UAP Perturbation strength ($\ell_1$ norm)  %learnt
by \lft and L2O  in various %training and evaluation 
scenarios ({\em lower is better}). The settings are consistent with  
%Figure\,\ref{fig: ASR_all}.
Figure~\ref{fig:rq2}
%\SL{Meta-test vs. meta-train? and missing baselines.}
%\SL{@Pu, update}
}
\label{fig: distortion_all}
%\vspace*{-0.2in}
\end{figure*}

\begin{figure*}[htb]
 \centering
   \begin{adjustbox}{max width=0.75\textwidth }
\begin{tabular}{m{1.0in}p{1.4in}p{1.4in}p{1.4in}}
\toprule
\diagbox[width=10em,trim=l]{{Meta-learning}}{{Meta-test}}
& \makecell{\centering \footnotesize{\texttt{T1}:  MNIST}}  
& \makecell{\centering \footnotesize{\texttt{T2}:  CIFAR-10}}
& \makecell{\centering  \footnotesize {\texttt{T3}: MNIST + CIFAR-10}}
\\
\midrule
\parbox[t][-1.2in][c]{0.4in}{ \centering \small{\texttt{L1}: MNIST} }%{\footnotesize (Congruous tasks)} }
 &   %\hspace*{-0.1in}
\includegraphics[width=1.5in]{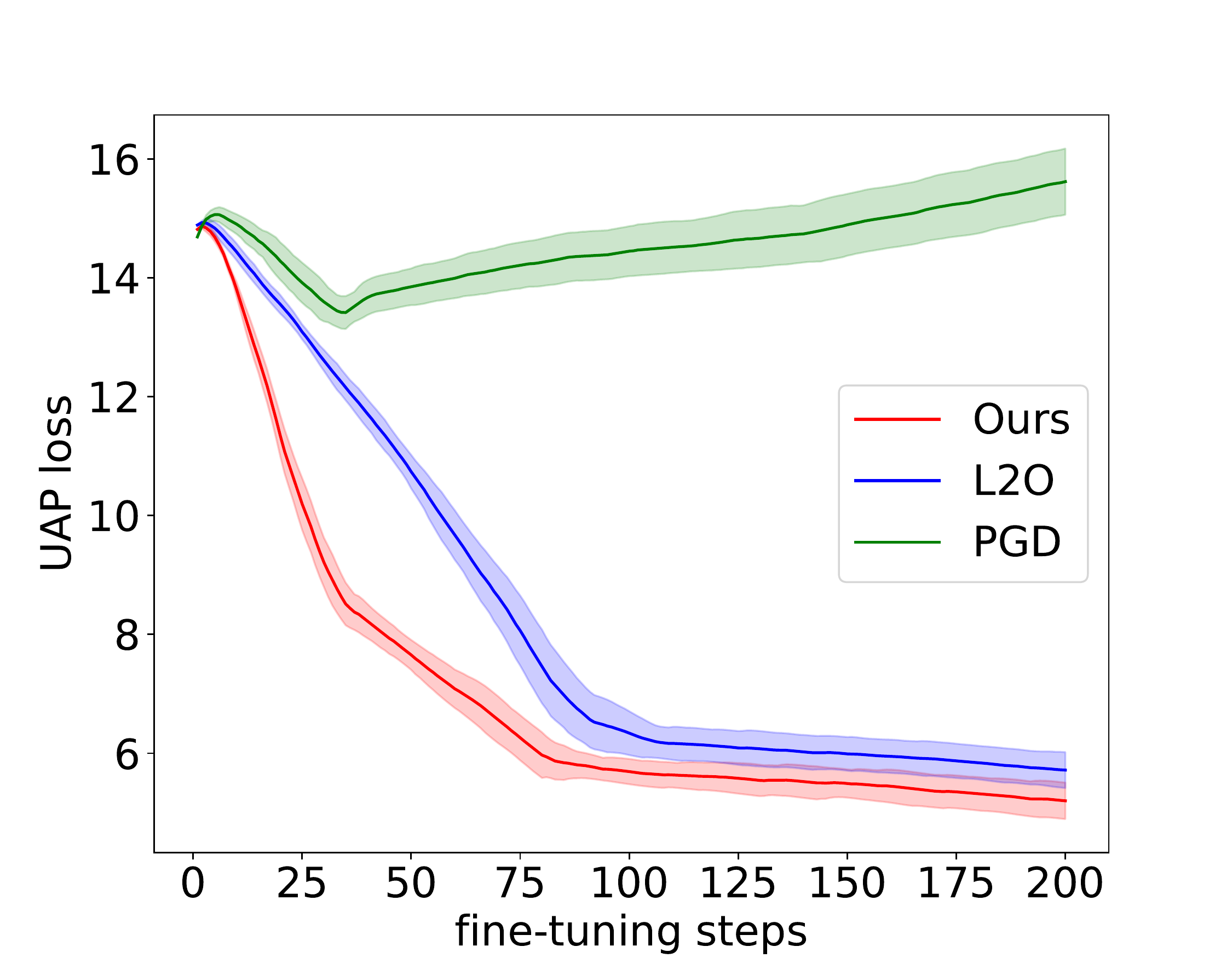}& %\hspace*{-0.1in}
\includegraphics[width=1.5in]{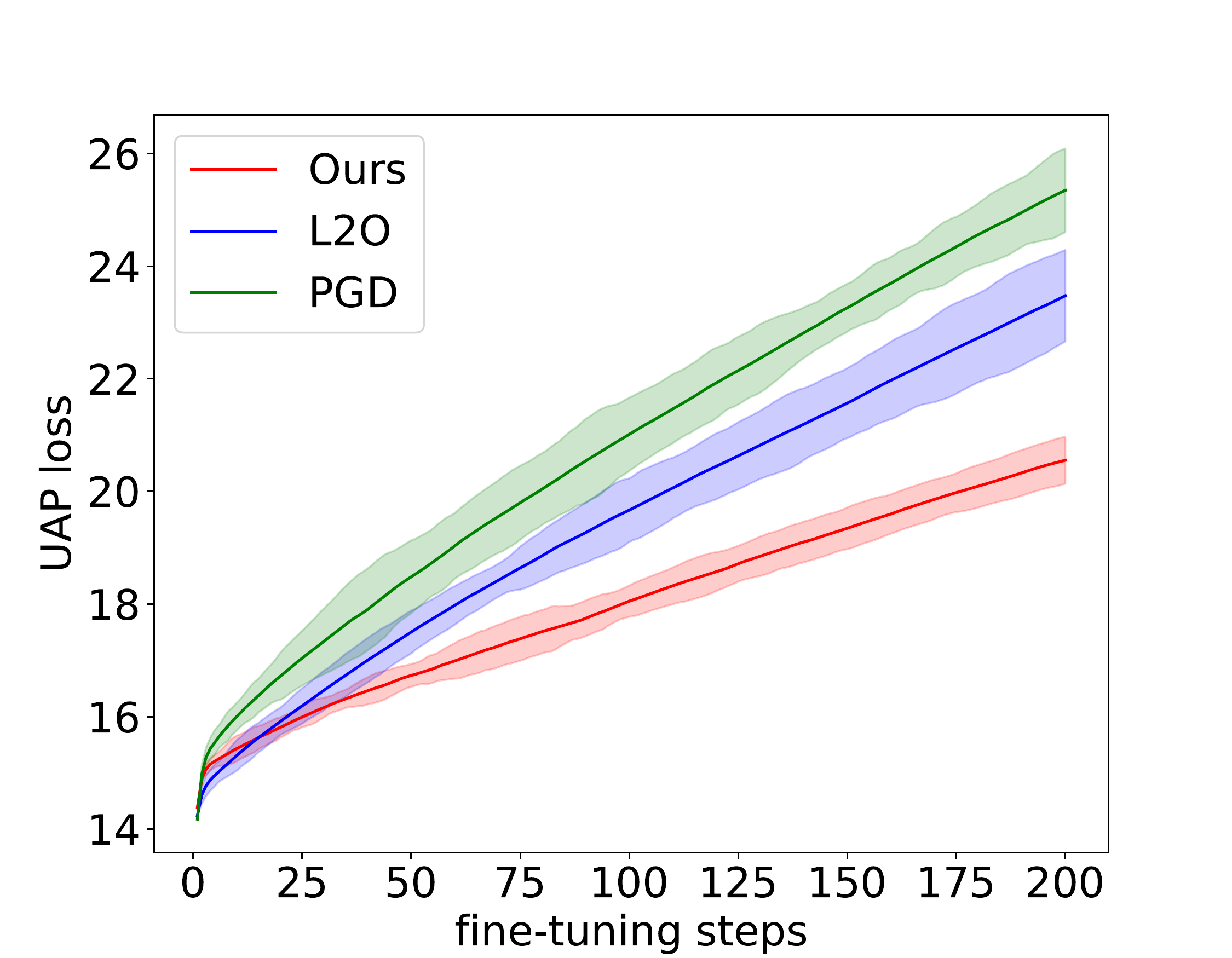}&
\includegraphics[width=1.5in]{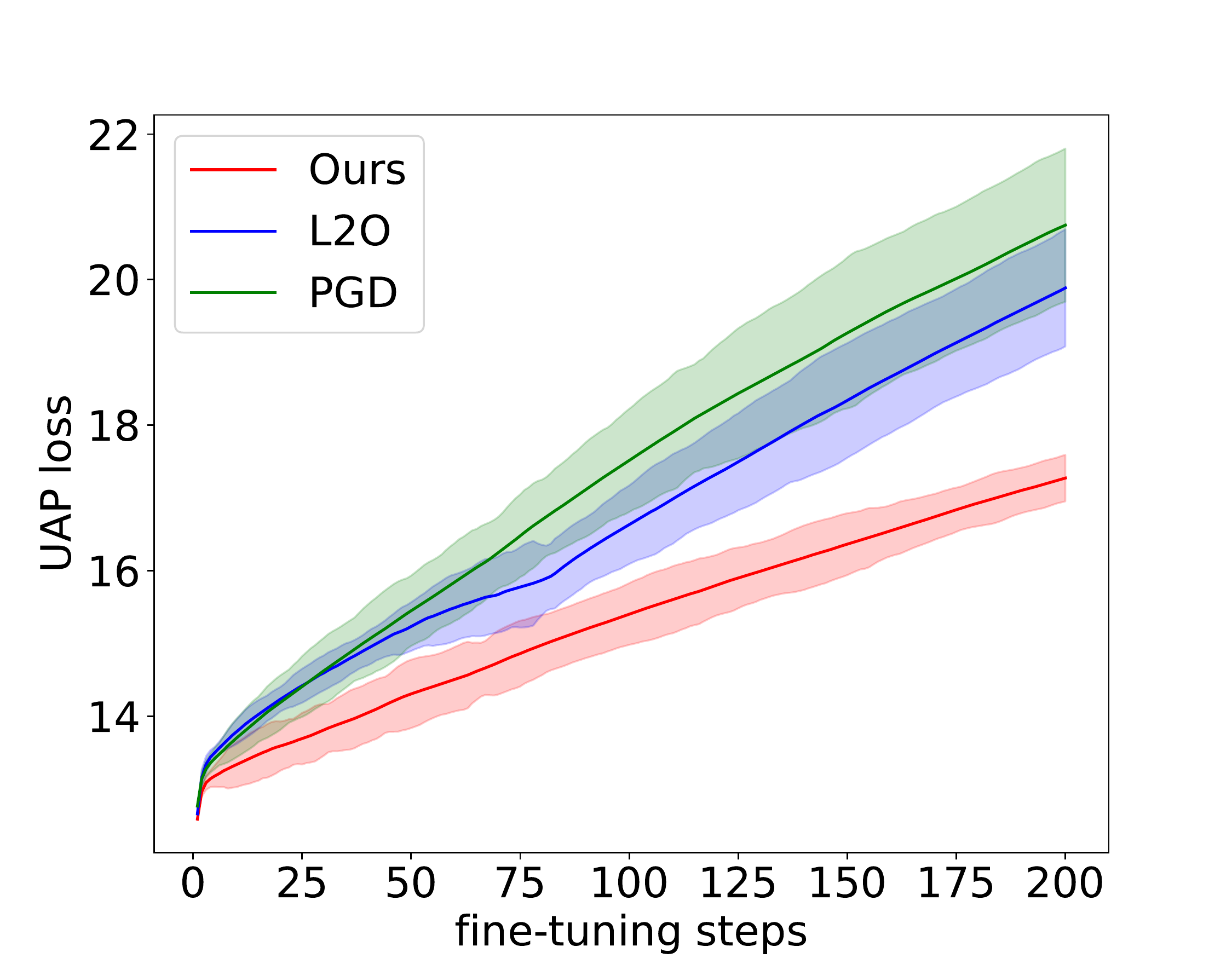}
\\ \midrule
%\makecell*[c]{ \centering trained on MNIST \\ \& CIFAR-10 }
\parbox[t][-1.2in][c]{0.4in}{ \centering \small{\texttt{L2}: MNIST + CIFAR-10}} %{\footnotesize (Incongruous tasks)}}
&   %\hspace*{-0.1in}
\includegraphics[width=1.5in]{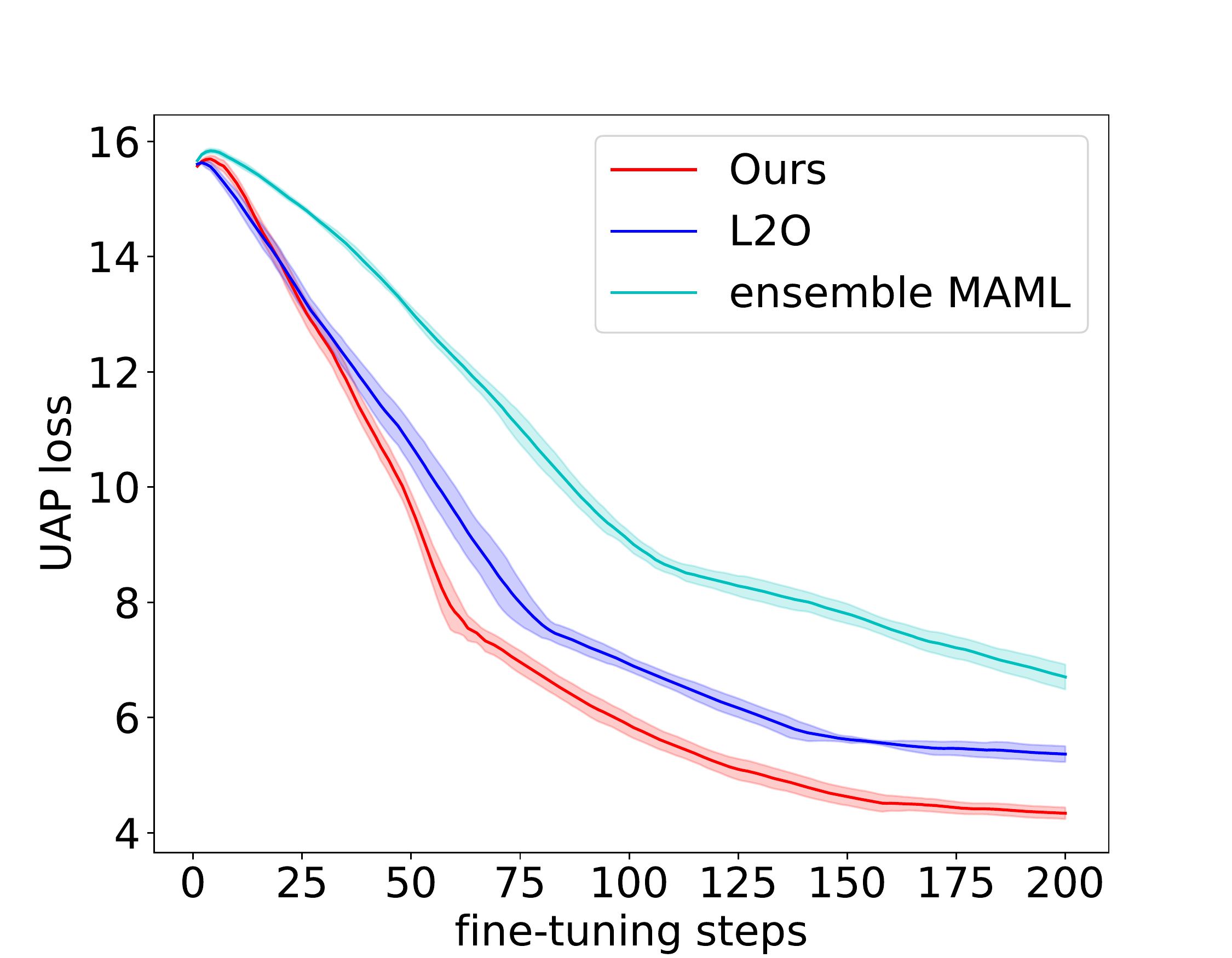} &  %\hspace*{-0.1in}
\includegraphics[width=1.5in]{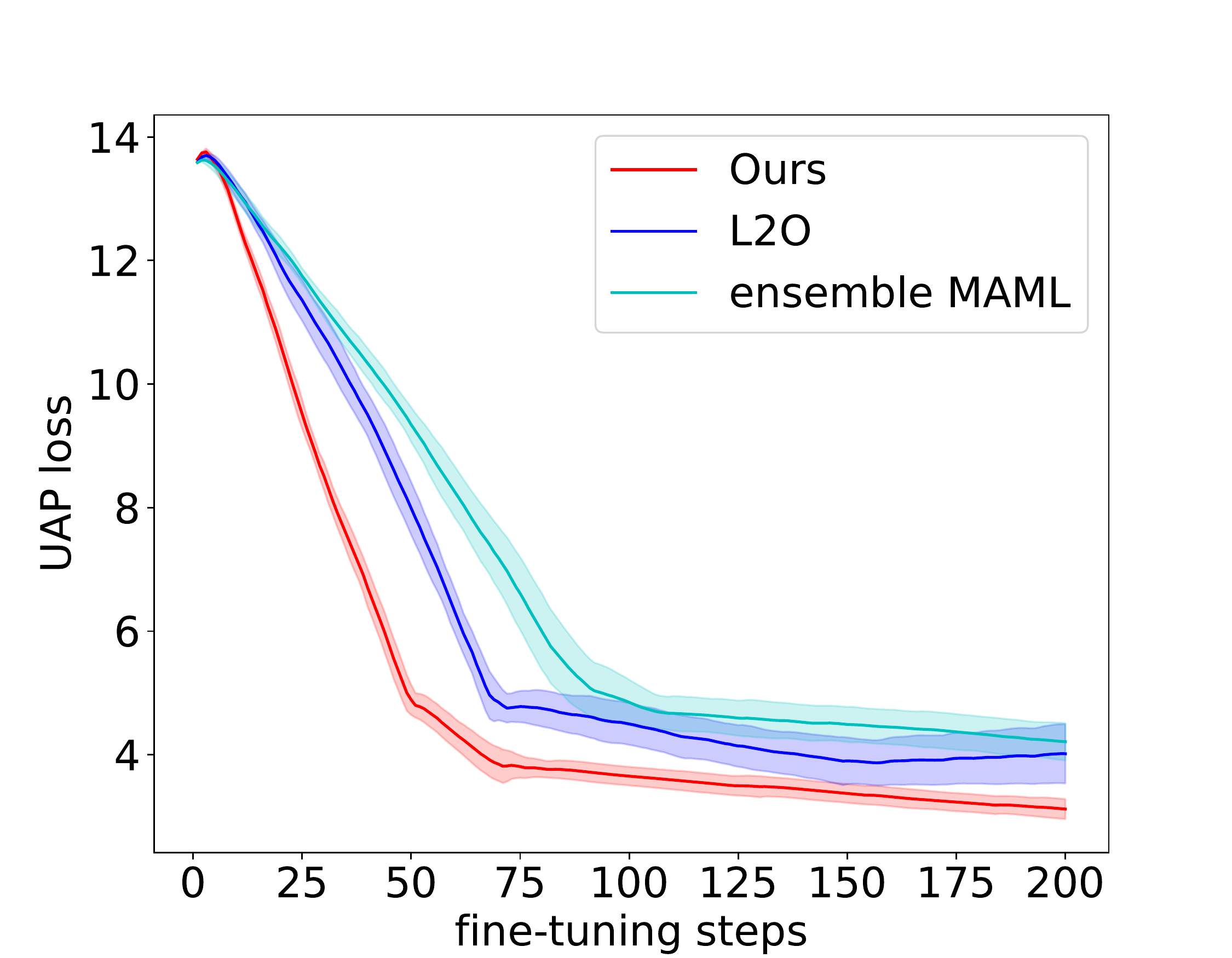} &
\includegraphics[width=1.5in]{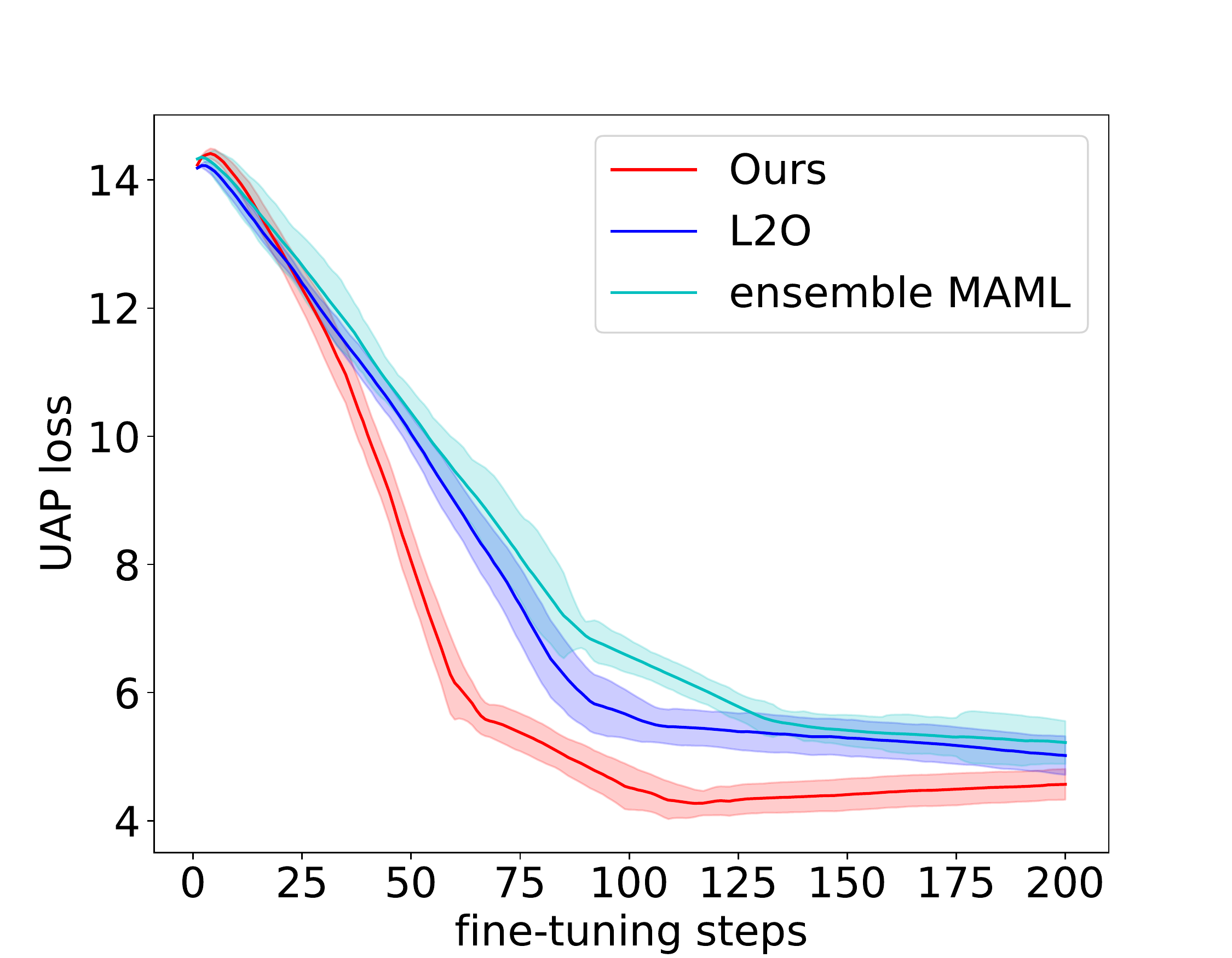}
\\
\bottomrule
\end{tabular}
 \end{adjustbox}
\caption{Fine-tuning loss of  UAP  by \lft and L2O in various %training and evaluation
scenarios ({\em lower is better}). The settings are consistent with 
%Figure\,\ref{fig: ASR_all}.
Figure~\ref{fig:rq2}.
%\SL{@Pu, update.}
%Figure 2 (in the main paper).
} 
\label{fig: loss_all}
%\vspace*{-0.2in}
\end{figure*}

\begin{figure*}[tb]
 \centering
 \begin{adjustbox}{max width=0.85\textwidth }
\begin{tabular}{m{0.6in}p{1.0in}p{1.0in}p{1.0in}p{1.0in}p{0.6in}p{0.1in}m{0.6in}p{1.0in}p{1.0in}p{1.0in}}
 \toprule
%\makecell*[c]{\centering trained on \\ MNIST }
\parbox[t][-0.9in][c]{0.6in}{\centering $\dout$ images }
 &   %\hspace*{-0.1in}
\includegraphics[width=1.0in]{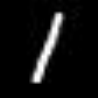}& 
\includegraphics[width=1.0in]{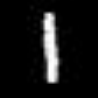}& 
\includegraphics[width=1.0in]{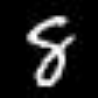}&
\includegraphics[width=1.0in]{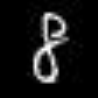} & 
\includegraphics[width=1.0in]{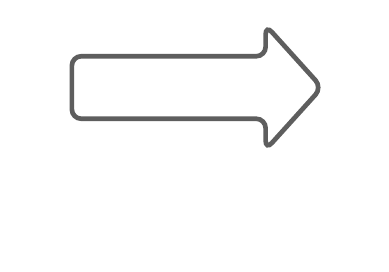}&  &
\parbox[t][-0.95in][c]{0.6in}{\centering UAP  } &    
\includegraphics[width=1.0in]{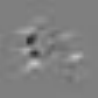} & 
\includegraphics[width=1.0in]{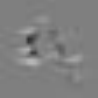} &  
\includegraphics[width=1.0in]{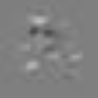} \\
  & & & & & & & & \makecell[c]{Ours} & \makecell[c]{L2O} 
& \makecell[c]{PGD} 
\\ \toprule
\parbox[t][-0.9in][c]{0.6in}{\centering $\dout$ images }
 &   %\hspace*{-0.1in}
\includegraphics[width=1.0in]{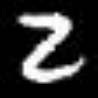}& 
\includegraphics[width=1.0in]{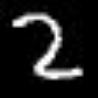}& 
\includegraphics[width=1.0in]{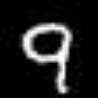}&
\includegraphics[width=1.0in]{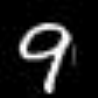} & 
\includegraphics[width=1.0in]{figs/Rarrow.pdf} &  & 
\parbox[t][-0.95in][c]{0.6in}{\centering UAP  } &  
\includegraphics[width=1.0in]{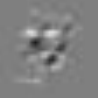} &  
\includegraphics[width=1.0in]{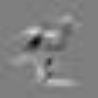} &   
\includegraphics[width=1.0in]{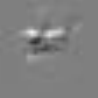} \\
&  &  && &  & & & \makecell[c]{Ours} & \makecell[c]{L2O} %& \makecell[c]{MAML}
& \makecell[c]{PGD} \\
\bottomrule
\end{tabular}
  \end{adjustbox} 
\caption{
Visualization of UAP patterns  by  \lft, L2O and PGD on MNIST for the same set of $\dout$ images.   
%The row `$\dout$ images' denotes  multiple images used as a validation set for all methods. The row `UAP' denotes  universal perturbation patterns derived by different optimizers ($2^{\text{nd}}$ and $4^{\text{th}}$ rows) for the same set of $\dout$ images ($1^{\text{st}}$ and $3^{\text{rd}}$ rows respectively). \SL{Update this figure.} \SL{This is out of place.}
}
\label{afig: perturbations}
\end{figure*}

Figure \ref{fig: distortion_all} shows
 $\ell_1$ norm of  UAP learnt by \lft and other baselines  in various training and evaluation scenarios. In general, the $\ell_1$ norm of the UAP learnt by \lft is smaller than that of using L2O or other baselines, except the case (MNIST, MNIST+CIFAR10). However, we recall from Figure\,\ref{fig:rq2} that the   ASR obtained by L2O  is much poorer than \lft in the case  (MNIST, MNIST+CIFAR10). 
Figure \ref{fig: loss_all} further demonstrates
the fine-tuning loss of the UAP learnt by \lft and other baselines. As expected, \lft yields a fast adaptation of UAP to attack unseen test images, corresponding to the lowest fine-tuning loss. If the ASR can hardly reach 100\%,  the loss does not decrease. 
We visualize the UAP patterns by LFT, L2O and PGD on MNIST for the same set of $\dout$ images in Figure~\ref{afig: perturbations}.

\section{Effect of the number of tasks in meta-learning} \label{appendix:num-tasks}

Here we study the effect of the number of tasks $N$ available for meta-learning on the meta-learning based schemes. For the experiments in \S \ref{sec:emp-eval}, we utilize $N = 1000$ for each meta-learning. We consider LFT as well as ensemble MAML as a baseline (as described in \S \ref{sec:emp-eval}) and meta-learn using tasks from both MNIST and CIFAR-10 (case \texttt{L2} in \S \ref{sec:emp-eval}). We consider $n = 100, 500, 1000$ tasks from each of the image sources and hence for meta-learning each of the MAMLs (by solving \eqref{eq:maml-orig}) for each image source. For meta-learning the LFT (with Algorithm~\ref{alg:LFT}) we will have $N = 2n = 200, 1000, 2000$ tasks since the same LFT can be meta-learned with tasks from multiple image sources.

We meta-test each of the schemes with 100 UAP generation tasks (obtained as described in \S \ref{sec:emp-eval}) from MNIST (case \texttt{T1} in \S \ref{sec:emp-eval}), CIFAR-10 (case \texttt{T2} in \S \ref{sec:emp-eval}) and MNIST+CIFAR (case \texttt{T3} in \S \ref{sec:emp-eval}). We present the ASR at 50 steps and 100 steps, and the number of steps needed to converge to 100\% ASR for each value of $N$ in Table~\ref{tab: ensemble_T}. As a reference point for non-meta-learning based schemes, the best average ASR achieved (i)~by S-UAP is 63\% for \texttt{T1}, 42\% for \texttt{T2} and 53\% for \texttt{T3}, and (ii)~by PGD is 50\% for \texttt{T1}, 25\% for \texttt{T2} and 38\% for \texttt{T3}. The results indicate that, for all values of $N$, our proposed LFT outperforms our ensemble MAML in terms of ASR (18-50\% higher ASR at 50 steps, 5-25\% higher ASR at 100 steps) as well as requires significantly smaller number of steps to converge to 100\% ASR. Overall, both methods improve the ASR with increasing $N$. However, even with just 100 tasks per image source (corresponding to $2 \text{ classes }\times 2 \text{ images per class } \times 2 \text{ sets for } \din_i \& \dout_i \times 100 = 800$ total images per source), LFT is able to achieve 100\% in almost all meta-test scenarios, demonstrating that LFT shows benefit even when the number of tasks available for meta-learning is relatively small. These numbers also show that, even with a small number of meta-learning tasks, LFT continues to outperform PGD and S-UAP baselines.

\begin{table}[ht]
\centering
\caption{ASR with the standard deviation of UAPs generated by  \lft and ensemble MAML, following the setting of Table~\ref{tab:rq2}. $N$ denotes the number of few-shot tasks for meta-learning. The performance is measured by (i)~highest ASR  within $50$ steps (ASR$_{50}$), (ii)~highest ASR  within $100$ steps (ASR$_{100}$), (iii)~step \# when first reaching $100\%$ ASR.
As a point of comparison, the best average ASR achieved (i)~by S-UAP is 63\% for \texttt{T1}, 42\% for \texttt{T2} and 53\% for \texttt{T3}, and (ii)~by PGD is 50\% for \texttt{T1}, 25\% for \texttt{T2} and 38\% for \texttt{T3}.
}
\label{tab: ensemble_T}
\scalebox{1.05}[1.0]{
\begin{adjustbox}{max width=0.85\textwidth }
 \begin{threeparttable}
\begin{tabular}{c|c|c|c|c|c|c|c|c|c|c}
\hline
\toprule[1pt]
\multicolumn{2}{c|}{\multirow{2}{*}{  \diagbox[width=10em,trim=l]{{Method }}{{Testing}}  }}
&  \multicolumn{3}{c|}{\texttt{T1}: MNIST}
&  \multicolumn{3}{c|}{\texttt{T2}: CIFAR-10}
&  \multicolumn{3}{c}{\texttt{T3}: MNIST + CIFAR-10}\\
\cline{3-11}
%\hline
%\midrule[1pt]
\multicolumn{2}{c|}{ } &
\makecell{ASR$_{50}$}  & \makecell{ASR$_{100}$}  & \makecell{step \#
%\tnote{1}
}
&
\makecell{ASR$_{50}$}  & \makecell{ASR$_{100}$} & \makecell{step \#}   &
\makecell{ASR$_{50}$}  & \makecell{ASR$_{100}$} & \makecell{step \#}   \\
%\hline
\midrule[1pt]
 \multirow {3}{*}{Ours } % & MAML  &  52\% & 0.14 & N/A & N/A & N/A & N/A & N/A  & N/A & N/A\\
%\hline
& $N=200$ &  50 $\pm$ 9   &  100 $\pm$ 2  & 98 & 68 $\pm$ 21 & 100 $\pm$ 0 & 79 & 31  $\pm$ 12 & 100  $\pm$ 0 & 92 \\
& $N=1000$ & 60 $\pm$ 17 &  100 $\pm$ 0 & 92  & 96 $\pm$ 7 &100  $\pm$ 0 & 55 & 86  $\pm$ 17 & 100  $\pm$ 0 & 68  \\
& $N=2000$ & 82 $\pm$ 8 &  100 $\pm$ 0  & 75 & 92 $\pm$ 11  & 100   $\pm$ 0 & 52 & 93 $\pm$ 5  & 100  $\pm$ 0& 58\\
 \midrule[1pt]
 \multirow {3}{*}{\makecell{ Ensemble \\ MAML }}
& $N=200$ & 0  $\pm$ 0 & 50$\pm$ 10  & 142 & 50 $\pm$ 0 & 92  $\pm$ 7 & 106 & 5  $\pm$ 9  & 82  $\pm$ 17& 120  \\
& $N=1000$ & 32 $\pm$ 15 &  75 $\pm$ 0 & 137  & 57 $\pm$ 18 & 96  $\pm$ 4& 88 & 54  $\pm$ 11 & 95  $\pm$ 6&  103 \\
& $N=2000$ & 50 $\pm$ 13 & 100$\pm$ 0  & 78 & 75 $\pm$ 16 & 100  $\pm$ 0 & 71 & 63  $\pm$ 7 & 100  $\pm$ 0 & 73\\
\bottomrule[1pt]
  \end{tabular}
\end{threeparttable}
\end{adjustbox}
}
\end{table}

\end{document}